\pdfoutput=1
\documentclass[sigconf, screen]{acmart}
\AtBeginDocument{%
  }

\copyrightyear{2023}
\acmYear{2023}
\setcopyright{cc}
\setcctype[4.0]{by}
\acmConference[CHI '23]{Proceedings of the 2023 CHI Conference on Human Factors in Computing Systems}{April 23--28, 2023}{Hamburg, Germany}
\acmBooktitle{Proceedings of the 2023 CHI Conference on Human Factors in Computing Systems (CHI '23), April 23--28, 2023, Hamburg, Germany}
\acmDOI{10.1145/3544548.3580816}
\acmISBN{978-1-4503-9421-5/23/04}

\usepackage{subcaption}
\usepackage{enumitem}
\usepackage{rotating}
\usepackage{microtype}
\usepackage{bm}
\usepackage{soul}
\usepackage{wrapfig}

\usepackage{balance}
\usepackage[varqu]{zi4}
\usepackage[utf8x]{inputenc}
\usepackage{units}
\usepackage{booktabs}
\usepackage{colortbl}

\usepackage{stfloats} %

\usepackage{amsmath}
\usepackage{mathtools}

\usepackage{acmart-taps}

\definecolor{orange}{RGB}{250,130,49}
\definecolor{red}{RGB}{234,59,90}
\definecolor{agreen}{RGB}{74, 198, 148}
\definecolor{purple}{RGB}{158, 62, 177}
\definecolor{darkpurple}{RGB}{170, 70, 210}
\definecolor{aqua}{RGB}{87, 180, 181}
\definecolor{lightblue}{RGB}{72, 123, 232}
\definecolor{hotpink}{RGB}{255, 83, 115}
\definecolor{teal}{RGB}{90, 200, 250}
\definecolor{linkColor}{RGB}{0, 128, 229}
\definecolor{lightgreen}{RGB}{33, 222, 128}
\definecolor{gray}{RGB}{75, 101, 132}

\definecolor{myred}{RGB}{224, 49, 119}
\definecolor{myorange}{RGB}{250, 130, 49}
\definecolor{myyellow}{RGB}{254, 211, 48}
\definecolor{mygreen}{RGB}{14, 152, 136}
\definecolor{myblue}{RGB}{0, 128, 229}
\definecolor{myviolet}{RGB}{56, 103, 214}
\definecolor{mypurple}{RGB}{136, 84, 208}
\definecolor{mybrown}{RGB}{132, 99, 88}
\definecolor{mygray}{RGB}{220, 220, 220}

\definecolor{tablegray}{RGB}{80, 80, 80}
\definecolor{refColor}{RGB}{109, 35, 130}

\newcommand{\link}[1]{{\href{#1}{\color{linkColor}\textbf{\texttt{#1}}}}}
\newcommand{\linkhref}[2]{{\href{#1}{\color{linkColor}\textbf{#2}}}}
\newcommand{\figpart}[1]{\textcolor{refColor}{#1}}

\newcommand{\tcolor}[2]{\textcolor{#1}{#2}}

\newcommand{\mypar}[1]{\vspace{3pt}\textbf{{#1}}}

\newcommand{\ra}[1]{\renewcommand{\arraystretch}{#1}}

\newcommand{\aci}[1]{\todo[linecolor=myorange,backgroundcolor=myorange!25,bordercolor=myorange]{1AC: ``#1''}}
\newcommand{\reaci}[2]{\todo[linecolor=myorange,backgroundcolor=myorange!25,bordercolor=myorange]{1AC: ``#1'' -- #2}}

\newcommand{\acii}[1]{\todo[linecolor=myblue,backgroundcolor=myblue!25,bordercolor=myblue]{2AC: ``#1''}}
\newcommand{\reacii}[2]{\todo[linecolor=myblue,backgroundcolor=myblue!25,bordercolor=myblue]{2AC: ``#1'' -- #2}}

\newcommand{\revieweri}[1]{\todo[linecolor=mygreen,backgroundcolor=mygreen!25,bordercolor=mygreen]{R2: ``#1''}}
\newcommand{\rerevieweri}[2]{\todo[linecolor=mygreen,backgroundcolor=mygreen!25,bordercolor=mygreen]{R2: ``#1'' -- #2}}

\newcommand{\reviewerii}[1]{\todo[linecolor=mypurple,backgroundcolor=mypurple!25,bordercolor=mypurple]{R3: ``#1''}}
\newcommand{\rereviewerii}[2]{\todo[linecolor=mypurple,backgroundcolor=mypurple!25,bordercolor=mypurple]{R3: ``#1'' -- #2}}

\newcommand{\reviewerclean}{
  \renewcommand{\aci}[1]{}
  \renewcommand{\reaci}[1]{}
  \renewcommand{\acii}[1]{}
  \renewcommand{\reacii}[1]{}
  \renewcommand{\revieweri}[1]{}
  \renewcommand{\rerevieweri}[1]{}
  \renewcommand{\reviewerii}[1]{}
  \renewcommand{\rereviewerii}[1]{}
}
\reviewerclean{}

\newcommand{\tool}{\textsc{\textsf{GAM Coach}}}

\newcommand{\menu}{\textit{Coach Menu}}
\newcommand{\panel}{\textit{Feature Panel}}
\newcommand{\mybookmark}{\textit{Bookmarks}}
\newcommand{\card}{\textit{Feature Card}}
\newcommand{\cards}{\textit{Feature Cards}}

\def\set#1{\{#1\}}

\definecolor{soulorange}{RGB}{255, 212, 153}
\definecolor{soulgray}{RGB}{220, 220, 220}
\colorlet{soulmyblue}{myblue!30}

\newcommand{\orangehl}[1]{{\sethlcolor{soulorange}\hl{#1}}}
\newcommand{\grayhl}[1]{{\sethlcolor{soulgray}\hl{#1}}}

\newcommand*{\vcenteredhbox}[1]{\begingroup\setbox0=\hbox{#1}\parbox{\wd0}{\box0}\endgroup}
\newcommand*\myquote[1]{``\textit{#1}''} 
\begin{document}

\title{\tool{}: Towards Interactive and User-centered Algorithmic Recourse}

\author{Zijie J. Wang}
\orcid{0000-0003-4360-1423}
\affiliation{%
  \institution{Georgia Tech}
  \city{Atlanta}
  \country{USA}
}

\author{Jennifer Wortman Vaughan}
\orcid{0000-0002-7807-2018}
\affiliation{%
  \institution{Microsoft Research}
  \city{New York}
  \country{USA}
}

\author{Rich Caruana}
\orcid{0000-0002-6383-7786}
\affiliation{%
  \institution{Microsoft Research}
  \city{Redmond}
  \country{USA}
}

\author{Duen Horng Chau}
\orcid{0000-0001-9824-3323}
\affiliation{%
  \institution{Georgia Tech}
  \city{Atlanta}
  \country{USA}
}

\renewcommand{\shortauthors}{Zijie J. Wang, et al.}

\begin{abstract}
Machine learning (ML) recourse techniques are increasingly used in high-stakes domains, providing end users with actions to alter ML predictions, but they assume ML developers understand what input variables can be changed.
However, a recourse plan's actionability is subjective and unlikely to match developers' expectations completely.
We present \tool{}, a novel open-source system that adapts integer linear programming to generate customizable counterfactual explanations for Generalized Additive Models (GAMs), and leverages interactive visualizations to enable end users to iteratively generate recourse plans meeting their needs.
A quantitative user study with 41 participants shows our tool is usable and useful, and users prefer personalized recourse plans over generic plans.
Through a log analysis, we explore how users discover satisfactory recourse plans, and provide empirical evidence that transparency can lead to more opportunities for everyday users to discover counterintuitive patterns in ML models.
\tool{} is available at: \link{https://poloclub.github.io/gam-coach/}.
\end{abstract}




\begin{teaserfigure}
  \centering
  \includegraphics[width=440pt]{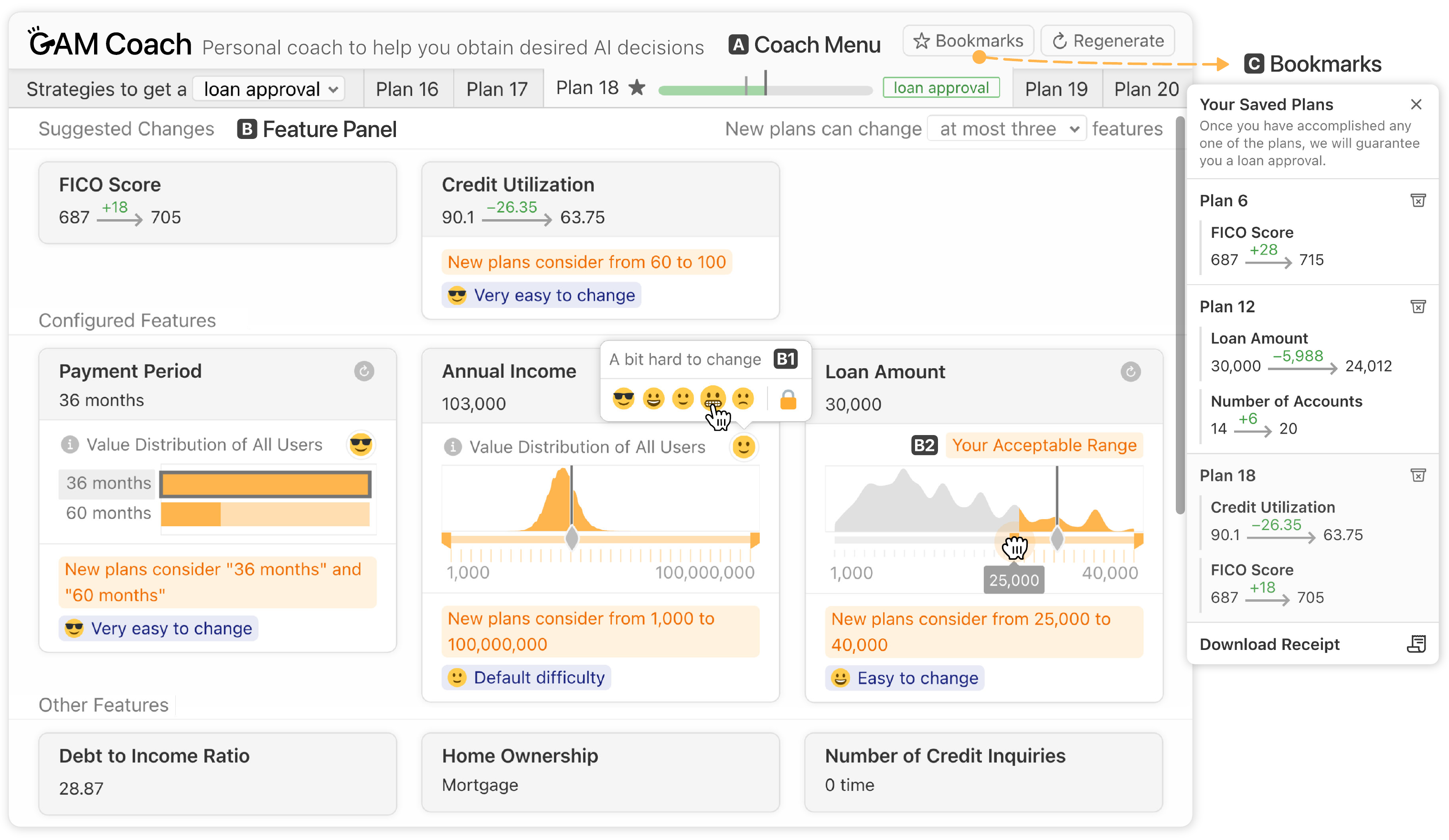}
  \vspace{-5pt}
  \caption[]{
    With a novel interface and an adaptation of integer linear programming, \tool{} empowers people impacted by machine learning-based decision-making systems to iteratively generate algorithmic recourse plans that reflect their preferences.
    Take loan application as an example.
    \textbf{(A)} The \menu{} helps a rejected loan applicant browse diverse recourse plans that would lead to loan approval.
    After the user selects a plan, \textbf{(B)} the \panel{} visualizes all feature information with progressive disclosure, enabling users to explore how hypothetical inputs affect the model's decision and specify recourse preferences---such as \textbf{(B1)} the \textit{difficulty} of changing a feature and \textbf{(B2)} its \textit{acceptable range} of values---guiding \tool{} to generate actionable plans.
    \textbf{(C)} The \mybookmark{} window allows users to compare bookmarked plans and save a verifiable receipt.
  }
  \vspace{3pt}
  \Description{
    A screenshot of GAM Coach user interface.
    On the top of the interface, there is a header including two buttons ``Bookmarks'' and ``Regenerate''.
    Below the header, there is a tab bar with five recourse plans.
    Below the tab bar, there is a feature panel with a grid of three feature cards.
    Feature cards are organized by ``suggested changes'', ``configured features'', and ``other features''.
    There are two types of feature cards.
    The continuous feature card contains a filled curve plot on top of a slider spanning the range of that variable.
    The categorical feature card contains a horizontal bar chart where each bar is one category.
    Users can drag the slider or click the bars to specify acceptable ranges
    Users can click an emoji button to specify the difficulty to change a feature.
    On the right of the main interface, there is a Bookmarks pop-up window.
    The window summarizes saved recourse plans and there is a ``download receipt'' button at the bottom of the window.'
  }
  \label{fig:teaser}
\end{teaserfigure}

\maketitle
\section{Introduction}

\begin{figure}[tb]
  \includegraphics[width=\linewidth]{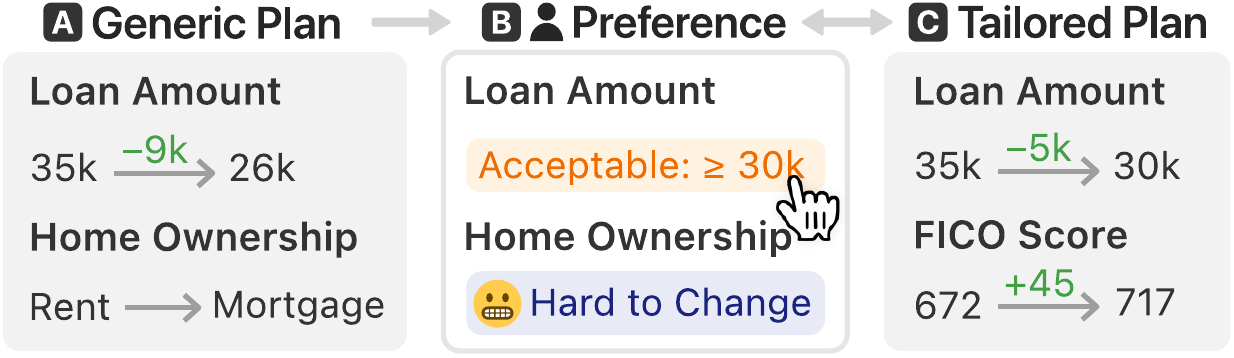}
  \caption[]{
    \textbf{\tool{} enables end users to iteratively fine-tune recourse plans.}
    \textbf{(A)} If a user finds the initial generic plan less actionable,
    \textbf{(B)} they can specify their recourse preferences through simple interactions.
    \textbf{(C)} Our tool will then generate tailored plans that reflect the user's preferences.
  }
  \Description{
    A flow chart of three components labeled A, B, and C.
    Component A shows a generic plan with loan amount decreasing from 35k to 26k and home ownership changing from rent to mortgage.
    Component B shows user preference that the acceptable range of loan amount is greater or equal to 30k and the difficulty of home ownership is hard to change.
    Component C shows a tailored plan with loan amount decreasing from 35k to 30k and FICO score increasing from 672 to 717.
    Component A points to component B.
    Component B and C point to each other.
  }
  \label{fig:crown}
\end{figure}

As machine learning (ML) is increasingly used in high-stakes decision-making, such as lending~\cite{siddiqiCreditRiskScorecards2013}, hiring~\cite{liemPsychologyMeetsMachine2018}, and college admissions~\cite{watersGRADEMachineLearning2014}, there has been a call for greater transparency and increased opportunities for algorithmic recourse~\cite{wachterCounterfactualExplanationsOpening2017}.
Algorithmic recourse aims to help those impacted by ML systems learn about the decision rules used~\cite{selbstIntuitiveAppealExplainable2018}, and provide suggestions for \textit{actions} to change decision outcome in the future~\cite{ustunActionableRecourseLinear2019}.
This often involves generating counterfactual (CF) examples,
which suggest minimal changes in a few features that would have led to the desired decision outcome~\cite{wachterCounterfactualExplanationsOpening2017}, such as ``if you had decreased your requested loan amount by \$9k and changed your home ownership from renting to mortgage, your loan application would have been approved.''~(\autoref{fig:crown}\figpart{A})

For such approaches to be useful, it is necessary for the suggested actions to be \textit{actionable}---realistic actions that users can appreciate and follow in their real-life circumstances. In the example above, changing home ownership status would arguably not be an actionable suggestion for most loan applicants.
To provide actionable recourse, recent work proposes techniques such as generating concise CF examples~\cite{leGRACEGeneratingConcise2020}, creating a diverse set of CF examples~\cite{mothilalExplainingMachineLearning2020,russellEfficientSearchDiverse2019}, and grouping features into different actionability categories~\cite{karimiAlgorithmicRecourseCounterfactual2021}.
These approaches often rely on the underlying assumption that ML developers can measure and predict which CF examples are actionable for all users.
However, the actionability of recourse is ultimately subjective and varies from one user to another~\cite{vermaCounterfactualExplanationsMachine2020,barocasHiddenAssumptionsCounterfactual2020}, or even for a single user at different times~\cite{zahediUnderstandingUserPreferences2019,lombrozoExplanatoryPreferencesShape2016}.
Therefore, there is a pressing need to capture and integrate user preferences into algorithmic recourse~\cite{kirfelWhatIfHow2021,barocasHiddenAssumptionsCounterfactual2020}.
\tool{} aims to take a user-centered approach~(\autoref{fig:crown}\figpart{B--C}) to fill this critical research gap.
In this work, we \textbf{contribute}: \looseness=-1

\aptLtoX[graphic=no,type=html]{
\begin{itemize}
  \item \textbf{\tool{}, the first interactive algorithmic recourse tool that empowers end users} to specify their recourse \textit{preferences}, such as difficulty and acceptable range for changing a feature, and iteratively \textit{fine-tune} actionable recourse plans~(\autoref{fig:teaser}).
  With an exploratory interface design~\cite{shneidermanBridgingGapEthics2020}, our tool helps users understand the ML model behaviors by experimenting with hypothetical input values and inspecting their effects on the model outcomes.
  Our tool advances over existing interactive ML tools~\cite{gomezViCEVisualCounterfactual2020,wexlerWhatIfToolInteractive2019}, overcoming unique design challenges identified from a literature review of recent algorithmic recourse work~(\autoref{sec:goal}, \autoref{sec:ui}).

  \item \textbf{Novel adaptation of integer linear programming to generate CF examples.}
  To operationalize interactive recourse, we ground our research in generalized additive models (GAMs)~\cite{nelderGeneralizedLinearModels1972,caruanaIntelligibleModelsHealthCare2015}, a popular class of models that performs competitively to other state-of-the-art models yet has a transparent and simple structure~\cite{wangPursuitInterpretableFair2020,changHowInterpretableTrustworthy2021, weldChallengeCraftingIntelligible2019,noriInterpretMLUnifiedFramework2019}.
  GAMs enable end users to probe model behaviors with hypothetical inputs in real time directly in web browsers.
  Adapting integer linear programming, we propose an efficient and flexible method to generate optimal CF examples for GAM-based classifiers and regressors with continuous and categorical features and pairwise feature interactions~\cite{louAccurateIntelligibleModels2013}~(\autoref{sec:method}).

  \item \textbf{Design lessons distilled from a user study with log analysis.}
  We conducted an online user study with 41 Amazon Mechanical Turk workers to evaluate \tool{} and investigate how everyday users would use an interactive algorithmic recourse tool.
  Through analyzing participants' interaction logs and subjective ratings in a hypothetical lending scenario, our study highlights that \tool{} is usable and useful, and users prefer personalized recourse plans over generic plans.
  We discuss the \textit{characteristics} of users' satisfactory recourse plans, \textit{approaches} users take to discover them, and \textit{design lessons} for future interactive recourse tools.
  We also provide empirical evidence that with transparency, everyday users can discover and are often puzzled by counterintuitive patterns in ML models~(\autoref{sec:user}).

  \item \textbf{An open-source, web-based implementation} that broadens people's access to developing and using interactive algorithmic recourse tools.
  We implement our CF generation method in both Python and JavaScript, enabling future researchers to use it on diverse platforms.
  We develop \tool{} with modern web technologies such as WebAssembly, so that anyone can access our tool using their web browsers without the need for installation or a dedicated backend server.
  We open-source\footnote{\tool{} code: \link{https://github.com/poloclub/gam-coach}} our CF generation library and \tool{} system with comprehensive documentation\footnote{\tool{} documentation: \link{https://poloclub.github.io/gam-coach/docs}}~(\autoref{sec:ui:implement}).
  For a demo video of \tool{}, visit \link{https://youtu.be/ubacP34H9XE}.

\end{itemize}
}{
\begin{itemize}[topsep=5pt, itemsep=0mm, parsep=1mm, leftmargin=9pt]
  \item \textbf{\tool{}, the first interactive algorithmic recourse tool that empowers end users} to specify their recourse \textit{preferences}, such as difficulty and acceptable range for changing a feature, and iteratively \textit{fine-tune} actionable recourse plans~(\autoref{fig:teaser}).
  With an exploratory interface design~\cite{shneidermanBridgingGapEthics2020}, our tool helps users understand the ML model behaviors by experimenting with hypothetical input values and inspecting their effects on the model outcomes.
  Our tool advances over existing interactive ML tools~\cite{gomezViCEVisualCounterfactual2020,wexlerWhatIfToolInteractive2019}, overcoming unique design challenges identified from a literature review of recent algorithmic recourse work~(\autoref{sec:goal}, \autoref{sec:ui}).

  \item \textbf{Novel adaptation of integer linear programming to generate CF examples.}
  To operationalize interactive recourse, we ground our research in generalized additive models (GAMs)~\cite{nelderGeneralizedLinearModels1972,caruanaIntelligibleModelsHealthCare2015}, a popular class of models that performs competitively to other state-of-the-art models yet has a transparent and simple structure~\cite{wangPursuitInterpretableFair2020,changHowInterpretableTrustworthy2021, weldChallengeCraftingIntelligible2019,noriInterpretMLUnifiedFramework2019}.
  GAMs enable end users to probe model behaviors with hypothetical inputs in real time directly in web browsers.
  Adapting integer linear programming, we propose an efficient and flexible method to generate optimal CF examples for GAM-based classifiers and regressors with continuous and categorical features and pairwise feature interactions~\cite{louAccurateIntelligibleModels2013}~(\autoref{sec:method}).

  \item \textbf{Design lessons distilled from a user study with log analysis.}
  We conducted an online user study with 41 Amazon Mechanical Turk workers to evaluate \tool{} and investigate how everyday users would use an interactive algorithmic recourse tool.
  Through analyzing participants' interaction logs and subjective ratings in a hypothetical lending scenario, our study highlights that \tool{} is usable and useful, and users prefer personalized recourse plans over generic plans.
  We discuss the \textit{characteristics} of users' satisfactory recourse plans, \textit{approaches} users take to discover them, and \textit{design lessons} for future interactive recourse tools.
  We also provide empirical evidence that with transparency, everyday users can discover and are often puzzled by counterintuitive patterns in ML models~(\autoref{sec:user}).

  \item \textbf{An open-source, web-based implementation} that broadens people's access to developing and using interactive algorithmic recourse tools.
  We implement our CF generation method in both Python and JavaScript, enabling future researchers to use it on diverse platforms.
  We develop \tool{} with modern web technologies such as WebAssembly, so that anyone can access our tool using their web browsers without the need for installation or a dedicated backend server.
  We open-source\footnote{\tool{} code: \link{https://github.com/poloclub/gam-coach}} our CF generation library and \tool{} system with comprehensive documentation\footnote{\tool{} documentation: \link{https://poloclub.github.io/gam-coach/docs}}~(\autoref{sec:ui:implement}).
  For a demo video of \tool{}, visit \link{https://youtu.be/ubacP34H9XE}.

\end{itemize}
}

\noindent To design and evaluate a prospective interface~\cite{shneidermanBridgingGapEthics2020} for interactive algorithmic recourse, we situate \tool{} in loan application scenarios.
However, we caution that adapting \tool{} for real lending settings would require further research with financial and legal experts as well as people who would be impacted by the system.
Our goal is for this work to serve as a foundation for the design of future user-centered recourse and interpretable ML tools.
\section{Related Work}

\subsection{Algorithmic Recourse}
\label{sec:related:recourse}

Algorithmic recourse aims to design techniques that provide those impacted by ML systems with actionable feedback about how to alter the outcome of ML models.
Popularized by~\citet{wachterCounterfactualExplanationsOpening2017}, researchers typically generate this actionable feedback by creating CF examples.
Here, a CF example represents a recourse plan that contains minimal changes to the original input but leads to a different model prediction~\cite{karimiSurveyAlgorithmicRecourse2021,ustunActionableRecourseLinear2019}.
For example, a bank that uses ML models to inform loan application decisions can provide a rejected loan applicant with a recourse plan that suggests the applicant increase their annual income by \$5k so that they can obtain a loan approval.
CF examples not only inform end users about the key features contributing to the decision, but also provide suggestions that end users can act on to obtain the desired outcome~\cite{ustunActionableRecourseLinear2019}.
Researchers have developed various methods to generate CF examples, such as casting it as an optimization problem~\cite[e.g.,][]{cuiOptimalActionExtraction2015, russellEfficientSearchDiverse2019, ustunActionableRecourseLinear2019, kanamoriDACEDistributionAwareCounterfactual2020, wachterCounterfactualExplanationsOpening2017, mohammadiScalingGuaranteesNearest2021},
searching through similar samples~\cite[e.g.,][]{goyalCounterfactualVisualExplanations2019, keaneGoodCounterfactualsWhere2020, delaneyInstancebasedCounterfactualExplanations2021, vanlooverenInterpretableCounterfactualExplanations2020, schleichGeCoQualityCounterfactual2021},
and developing generative models~\cite[e.g.,][]{kennyGeneratingPlausibleCounterfactual2021, dhurandharExplanationsBasedMissing2018, joshiRealisticIndividualRecourse2019, singlaExplanationProgressiveExaggeration2020}.

It is challenging to generate helpful CF examples in practice.
Besides making minimal changes, a helpful CF example should also be \textit{actionable} for the end user~\cite{ustunActionableRecourseLinear2019, keaneIfOnlyWe2021}.
To generate actionable recourse plans, recent research includes proposals to find concise CF examples~\cite{leGRACEGeneratingConcise2020}, consider causality~\cite{karimiAlgorithmicRecourseCounterfactual2021,mahajanPreservingCausalConstraints2020, karimiAlgorithmicRecourseImperfect2020}, present diverse plans~\cite{mothilalExplainingMachineLearning2020,russellEfficientSearchDiverse2019}, and assign features with different actionability scores~\cite{karimiAlgorithmicRecourseCounterfactual2021}.
However, the actionability of recourse is ultimately subjective and varies among end users~\cite{vermaCounterfactualExplanationsMachine2020,kirfelWhatIfHow2021,zahediUnderstandingUserPreferences2019,lombrozoExplanatoryPreferencesShape2016}.
To restore users' autonomy with CF examples, some researchers explore the potential of interactive tools.
For example, Prospector~\cite{krauseInteractingPredictionsVisual2016}, What-If Tool~\cite{wexlerWhatIfToolInteractive2019}, Polyjuice~\cite{wuPolyjuiceGeneratingCounterfactuals2021}, and AdViCE~\cite{gomezAdViCEAggregatedVisual2021} leverage interactive visualizations to help ML developers debug models with CF examples.
Context Sight~\cite{yuanContextSightModel2022} allows ML developers to analyze model errors by customizing the acceptable feature range and desired number of changes in CF examples.
CEB~\cite{myersRevealingNeuralNetwork2020} interactively presents CF examples to help non-experts understand neural networks.
In comparison, \tool{} aims to empower \textit{end users} to discover \textit{actionable} strategies to alter undesirable ML decisions.

DECE~\cite{chengDECEDecisionExplorer2021} is a visual analytics tool designed to help ML developers and end users interpret neural network predictions with CF examples.
It allows users to customize CF examples by specifying acceptable feature ranges.
In comparison, while the interface for \tool{} is model agnostic, the recourse generation technique it employs is tailored to GAMs, a different model family, and our tool especially focuses on end users without an ML background.
We evaluate \tool{} through an observational log study with 41 crowdworkers, while DECE is evaluated through three expert interviews.
These evaluations provide complementary viewpoints and insights into how interactive recourse tools may be used in practice.
Possibly closest in spirit to our work is ViCE~\cite{gomezViCEVisualCounterfactual2020}, an interactive visualization tool that generates CF examples on end users' selected continuous features.
In contrast, \tool{}---which supports both continuous and categorical features, as well as their pairwise interactions---allows end users to specify a much wider range of recourse preferences including feature difficulty, acceptable range, and the number of features to change. Our tool then generates \textit{optimal} and \textit{diverse} CF examples meeting specified preferences.

\subsection{Interactive Tools for Interpretable ML}

Besides CF explanations, researchers have developed interactive tools to help different ML stakeholders interpret ML models~\cite[e.g.,][]{wangTimberTrekExploringCurating2022, hohmanSUMMITScalingDeep2019,kahngActiVisVisualExploration2018,pezzottiDeepEyesProgressiveVisual2018}.
In particular, the simple structure and high performance of GAMs have attracted many researchers to use this model to explore how interactivity plays a role in interpretable ML.
For example, Gamut~\cite{hohmanGamutDesignProbe2019} provides both global and local explanations by visualizing the shape functions in GAMs.
Similarly, TeleGam~\cite{hohmanTeleGamCombiningVisualization2019} helps users understand GAM predictions by combining both graphical and textual explanations.
GAM Changer~\cite{wangInterpretabilityThenWhat2022} supports users to edit GAM model parameters through interactive visualization.
However, the target users of these tools are ML experts, such as ML researchers and model developers, or domain experts who need to vet and correct models before deployment.
In comparison, \tool{} targets people who are impacted by ML models and who are less knowledgeable about ML and domain-specific concepts~\cite{sureshExpertiseRolesFramework2021}.

There is an increasing body of research in developing interactive systems to help \textit{non-experts} interact with ML models.
The main goal of these tools is to educate non-experts about the underlying mechanisms of ML models.
For example, Teachable Machine~\cite{carneyTeachableMachineApproachable2020} helps users learn about basic ML concepts through interactive demos.
Tensorflow Playground~\cite{smilkovDirectManipulationVisualizationDeep2017}, GAN Lab~\cite{kahngGANLabUnderstanding2019}, and CNN Explainer~\cite{wangCNNExplainerLearning2020} use interactive visualizations to help novices learn about the underlying mechanisms of neural networks, generative adversarial networks, and convolutional neural networks, respectively.
In contrast, instead of educating non-experts on the technical inner workings of ML models, we focus on helping non-experts who are impacted by ML models understand why a model makes a particular decision and what actions they can take to alter that decision.
\section{Design Goals}
\label{sec:goal}

Our goal is to design and develop an interactive, visual experimentation tool that respects end users' autonomy in algorithmic recourse, helping them discover and fine-tune recourse plans that reflect their preferences and needs.
We identify five main design goals of \tool{} through synthesizing the trends and limitations of traditional algorithmic recourse systems~\cite[e.g.,][]{barocasHiddenAssumptionsCounterfactual2020,karimiSurveyAlgorithmicRecourse2021,keaneIfOnlyWe2021,mittelstadtExplainingExplanationsAI2019,shneidermanBridgingGapEthics2020, wachterCounterfactualExplanationsOpening2017, abdulTrendsTrajectoriesExplainable2018}.

\aptLtoX[graphic=no,type=html]{
\begin{itemize}
  \leftskip-7pt
  \item [\textbf{G1.}]
  \textbf{Visual summary of diverse algorithmic recourse plans.}
  To help end users find actionable recourse plans, researchers suggest presenting diverse CF options that users can pick from~\cite{mothilalExplainingMachineLearning2020,barocasHiddenAssumptionsCounterfactual2020}.
  Thus, \tool{} should efficiently generate diverse recourse plans~(\autoref{sec:method:ip}) and present a visual summary of each plan as well as display multiple plans at the same time~(\autoref{sec:ui:tab}).
  This could help users compare different strategies and inform interactions to generate better recourse plans.

  \item [\textbf{G2.}]
  \textbf{Easy ways to specify recourse preferences.}
  What makes a recourse plan actionable varies from one user to another---it is crucial for a recourse tool to enable users to specify a wide range of recourse preferences~\cite{barocasHiddenAssumptionsCounterfactual2020,mittelstadtExplainingExplanationsAI2019,kirfelWhatIfHow2021}.
  Therefore, we would like to allow users to easily configure (1) the \textit{difficulty} of changing a feature, (2) the \textit{acceptable range} within which a feature can change, and (2) the \textit{maximum number of features} that a recourse plan can change~(\autoref{sec:ui:panel}), and \tool{} should generate plans reflecting users' specified preferences~(\autoref{sec:method:customization}).
  This interactive recourse design would empower users to iteratively customize recourse plans until they find satisfactory plans.

  \item [\textbf{G3.}]
  \textbf{Exploratory interface to experiment with hypothetical inputs.}
  The goal of algorithmic recourse is not only to help users identify actions to alter unfavorable model decisions, but also to help them understand how a model makes decisions~\cite{wachterCounterfactualExplanationsOpening2017,karimiSurveyAlgorithmicRecourse2021}.
  When explaining a model's decision-making, research shows that interfaces allowing users to probe an ML model with different inputs help users understand model behaviors and lead to greater satisfaction with the model~\cite{nourashrafeddinVisualApproachInteractive2018, chengExplainingDecisionMakingAlgorithms2019,shneidermanBridgingGapEthics2020,wexlerWhatIfToolInteractive2019}.
  Therefore, we would like \tool{} to enable users to experiment with different hypothetical inputs and inspect how these changes affect the model's decision~(\autoref{sec:ui:panel}).

  \item [\textbf{G4.}]
  \textbf{Clear communication and engagement.}
  The target users of \tool{} are everyday people who are usually less knowledgeable about ML and domain-specific concepts~\cite{sureshExpertiseRolesFramework2021}.
  Our goal is to design and develop an interactive system that is easy to understand and engaging to use, requiring the tool to communicate and explain recourse plans and domain-specific information to end users~(\autoref{sec:ui:panel}, \autoref{sec:ui:bookmark}).

  \item [\textbf{G5.}]
  \textbf{Open-source and model-agnostic implementation.}
  We aim to develop an interactive recourse tool that is easily accessible to users, with no installation required.
  By using web browsers as the platform, users can directly access \tool{} through their laptops or tablets.
  Additionally, we aim to make our interface model-agnostic so that future researchers can use it with different ML models and recourse techniques.
  Finally, we would like to open-source our implementation and provide documentation to support future design, research, and development of interactive algorithmic recourse~(\autoref{sec:ui:implement}).
\end{itemize}
}{
\begin{enumerate}[topsep=1mm, itemsep=0mm, parsep=1mm, leftmargin=18pt, label=\textbf{G\arabic*.}, ref=G\arabic*]
  \item \label{item:g1}
  \textbf{Visual summary of diverse algorithmic recourse plans.}
  To help end users find actionable recourse plans, researchers suggest presenting diverse CF options that users can pick from~\cite{mothilalExplainingMachineLearning2020,barocasHiddenAssumptionsCounterfactual2020}.
  Thus, \tool{} should efficiently generate diverse recourse plans~(\autoref{sec:method:ip}) and present a visual summary of each plan as well as display multiple plans at the same time~(\autoref{sec:ui:tab}).
  This could help users compare different strategies and inform interactions to generate better recourse plans.
  \item \label{item:g2}
  \textbf{Easy ways to specify recourse preferences.}
  What makes a recourse plan actionable varies from one user to another---it is crucial for a recourse tool to enable users to specify a wide range of recourse preferences~\cite{barocasHiddenAssumptionsCounterfactual2020,mittelstadtExplainingExplanationsAI2019,kirfelWhatIfHow2021}.
  Therefore, we would like to allow users to easily configure (1) the \textit{difficulty} of changing a feature, (2) the \textit{acceptable range} within which a feature can change, and (2) the \textit{maximum number of features} that a recourse plan can change~(\autoref{sec:ui:panel}), and \tool{} should generate plans reflecting users' specified preferences~(\autoref{sec:method:customization}).
  This interactive recourse design would empower users to iteratively customize recourse plans until they find satisfactory plans.

  \item \label{item:g3}
  \textbf{Exploratory interface to experiment with hypothetical inputs.}
  The goal of algorithmic recourse is not only to help users identify actions to alter unfavorable model decisions, but also to help them understand how a model makes decisions~\cite{wachterCounterfactualExplanationsOpening2017,karimiSurveyAlgorithmicRecourse2021}.
  When explaining a model's decision-making, research shows that interfaces allowing users to probe an ML model with different inputs help users understand model behaviors and lead to greater satisfaction with the model~\cite{nourashrafeddinVisualApproachInteractive2018, chengExplainingDecisionMakingAlgorithms2019,shneidermanBridgingGapEthics2020,wexlerWhatIfToolInteractive2019}.
  Therefore, we would like \tool{} to enable users to experiment with different hypothetical inputs and inspect how these changes affect the model's decision~(\autoref{sec:ui:panel}).

  \item \label{item:g4}
  \textbf{Clear communication and engagement.}
  The target users of \tool{} are everyday people who are usually less knowledgeable about ML and domain-specific concepts~\cite{sureshExpertiseRolesFramework2021}.
  Our goal is to design and develop an interactive system that is easy to understand and engaging to use, requiring the tool to communicate and explain recourse plans and domain-specific information to end users~(\autoref{sec:ui:panel}, \autoref{sec:ui:bookmark}).

  \item \label{item:g5}
  \textbf{Open-source and model-agnostic implementation.}
  We aim to develop an interactive recourse tool that is easily accessible to users, with no installation required.
  By using web browsers as the platform, users can directly access \tool{} through their laptops or tablets.
  Additionally, we aim to make our interface model-agnostic so that future researchers can use it with different ML models and recourse techniques.
  Finally, we would like to open-source our implementation and provide documentation to support future design, research, and development of interactive algorithmic recourse~(\autoref{sec:ui:implement}).
\end{enumerate}
}

\section{Techniques for Customizable Recourse Generation}
\label{sec:method}

Given our design goals~(\aptLtoX[graphic=no,type=html]{\textbf{G1}--\textbf{G5}}{\ref{item:g1}--\ref{item:g5}}), it is crucial for \tool{} to generate customizable recourse plans interactively with a short response time.
Therefore, we base our design on GAMs, a family of ML models that perform competitively to state-of-the-art models yet have a transparent and simple structure---enabling end users to probe model behaviors in real-time with hypothetical inputs.
In addition, with a novel adaptation of integer linear programming~(\autoref{sec:method:ip}), GAMs allow us to efficiently generate recourse plans that respect users' preferences and thus achieve our design goals~(\autoref{sec:method:customization}).

\subsection{Model Choice}
\label{sec:method:model}

To operationalize our design of interactive algorithmic recourse, we ground our research in GAMs~\cite{hastieGeneralizedAdditiveModels1999}. More specifically, we make use of a type of GAMs called \textit{Explainable Boosting Machines},  (EBMs)~\cite{caruanaIntelligibleModelsHealthCare2015,noriInterpretMLUnifiedFramework2019}, which perform competitively to the state-of-the-art black-box models yet have a transparent and simple structure~\cite{wangPursuitInterpretableFair2020,changHowInterpretableTrustworthy2021, weldChallengeCraftingIntelligible2019,noriInterpretMLUnifiedFramework2019}.
Compared to simple models like linear models or decision trees, EBMs achieve superior accuracy by learning complex relations between features through gradient-boosting trees~\cite{louAccurateIntelligibleModels2013}, and thus deploying our design is realistic.
Compared to complex models like neural networks, EBMs have a similar performance on tabular data but a simpler structure; therefore, users can probe model behaviors in real-time with hypothetical inputs~(\aptLtoX[graphic=no,type=html]{\textbf{G3}}{\ref{item:g3}}).

Given an \tcolor{myorange}{input} $\textcolor[HTML]{FA8231}{x \in \mathbb{R}^{k}}$ with \tcolor{myorange}{$k$ features}, the \tcolor{myred}{output} $\textcolor[HTML]{E03177}{y \in \mathbb{R}}$ of an EBM model can be written as:
\begin{align}
\begin{split}
    \label{equation:gam}
    \textcolor[HTML]{E03177}{y} &= \textcolor[HTML]{4B6584}{l \left( \textcolor[HTML]{E03177}{S_x} \right)} \\
    \textcolor[HTML]{E03177}{S_x} &= \textcolor[HTML]{0E9888}{\beta_0} + \textcolor[HTML]{0080E5}{f_1 \left(\textcolor[HTML]{FA8231}{x_1}\right)} + \textcolor[HTML]{0080E5}{f_2 \left(\textcolor[HTML]{FA8231}{x_2}\right)} + \cdots + \textcolor[HTML]{0080E5}{f_k \left(\textcolor[HTML]{FA8231}{x_k}\right)} + \cdots + \textcolor[HTML]{0080E5}{f_{ij}(\textcolor[HTML]{FA8231}{x_i, x_j})}
\end{split}
\end{align}

\noindent Here, each \tcolor{myblue}{shape function} $\textcolor[HTML]{0080E5}{f_j}$ for single features $j \in \set{1, 2, \dots, k}$ or $\textcolor[HTML]{0080E5}{f_{ij}(\textcolor[HTML]{FA8231}{x_i, x_j})}$ for pairwise interactions between features~\cite{louAccurateIntelligibleModels2013}
is learned using \tcolor{myblue}{gradient-boosted trees}~\cite{louIntelligibleModelsClassification2012}.
$\textcolor[HTML]{E03177}{S_x}$ is the sum of all \tcolor{myblue}{shape function} outputs as well as \tcolor{mygreen}{the intercept constant} $\textcolor[HTML]{0E9888}{\beta_0}$.
The model converts $\textcolor[HTML]{E03177}{S_x}$ to the \tcolor{myred}{output} $\textcolor[HTML]{E03177}{y}$ through a \tcolor{gray}{link function} $\textcolor[HTML]{4B6584}{l}$ that is determined by the ML task.
For example, a \tcolor{gray}{sigmoid function} is used for binary classifications, and an \tcolor{gray}{identity function} for regressions.

What distinguishes EBMs from other GAMs is that the \tcolor{myblue}{shape function~$f_j$ or $f_{ij}$} is an ensemble of trees, mapping a \tcolor{myorange}{main effect feature value $x_j$} or a \tcolor{myorange}{pairwise interaction $(x_i, x_j)$} to a scalar \tcolor{myblue}{score}.
Before training, EBM applies \textit{equal-frequency binning} on each continuous feature, where bins have different widths but the same number of training samples.
This discrete bucketing process is commonly used to speed up gradient-boosting tree methods with little cost in accuracy, such as in popular tree-based models LightGBM~\cite{keLightGBMHighlyEfficient2017} and XGBoost~\cite{chenXGBoostScalableTree2016}.
For categorical features, EBMs treat each discrete level as a bin.
Once an EBM model is trained, the learned parameters for each ensemble of trees which defines the feature split points and scores in each region defined by these split points are transformed to a \textit{lookup histogram} (for univariate features) and a \textit{lookup table} (for pairwise interactions).
When predicting on a data point, the model first looks up corresponding scores for all feature values and interaction terms and then applies \autoref{equation:gam} to compute the output. \looseness=-1

\subsection{\mbox{CF Generation: Integer Linear Programming}}
\label{sec:method:ip}

A recourse plan is a CF example $c$ that makes minimal changes to the original input $x$ but leads to a different prediction.
Without loss of generality, we use binary classification as an example, with \tcolor{gray}{sigmoid function} $\textcolor[HTML]{4B6584}{\sigma(a) = \frac{1}{1 + e^{-a}}}$ as a \tcolor{gray}{link function}.
If $\textcolor[HTML]{4B6584}{\sigma\left(\textcolor[HTML]{E03177}{S_x}\right)} \geq 0.5$ or $\textcolor[HTML]{E03177}{S_x} \geq 0$, the model predicts the input $x$ as positive; otherwise it predicts $x$ as negative.
To generate $c$, we can change $x$ so that the new score $\textcolor[HTML]{E03177}{S_c}$ has a different sign from $\textcolor[HTML]{E03177}{S_x}$.
Note that $\textcolor[HTML]{E03177}{S_x}$ is a linear combination of shape function scores and so is $\textcolor[HTML]{E03177}{S_c - S_x}$.
Thus, we can express this counterfactual constraint as a linear constraint~(derivation in \autoref{sec:milp:cf}).
To enforce $c$ to only make minimal changes to $x$, we can minimize the distance between $c$ and $x$, which can also be expressed as a linear function~(\autoref{sec:milp:proximity}).
Since all constraints are linear, and there are a finite number of bins for each feature, we express the \tool{} recourse generation as an \textit{integer linear program}:
\aptLtoX[graphic=no,type=html]{\renewcommand\theequation{2\alph{equation}}
\setcounter{equation}{0}\begin{eqnarray}
&&    \min \phantom{.}  \textnormal{distance} \\
&&\quad    \textnormal{s.t.} \phantom{.}  \textnormal{distance} = \sum_{i=1}^{k} \sum_{b\in{B_i}} d_{ib} \textcolor[HTML]{FA8231}{v_{ib}}  \label{ilp:distance} \\
&&\qquad \textcolor[HTML]{E03177}{-S_x} \leq \sum_{i=1}^{k} \sum_{b\in{B_i}} g_{ib} \textcolor[HTML]{FA8231}{v_{ib}} + \sum_{\left(i, j\right) \in N} \sum_{b_1 \in B_i} \sum_{b_2 \in B_j} h_{ijb_1b_2} \textcolor[HTML]{0E9888}{z_{ijb_1b_2}}  \label{ilp:cf} \\
&&\qquad \textcolor[HTML]{0E9888}{z_{ijb_1b_2}} = \textcolor[HTML]{FA8231}{v_{ib_1} v_{jb_2}}\quad  \textnormal{for } \left(i, j\right) \in N, \enskip b_1 \in B_i, \enskip b_2 \in B_j  \label{ilp:interaction}\\
&&\qquad \sum_{b\in{B_i}}^{} \textcolor[HTML]{FA8231}{v_{ib}} \leq 1\quad \textnormal{for } i = 1, \dots, k  \label{ilp:one}\\
&&\qquad \textcolor[HTML]{FA8231}{v_{ib}} \in \left\{0, 1\right\}\quad \textnormal{for } i = 1, \dots, k, \enskip b \in B_i   \label{ilp:binaryv}\\
&&\qquad \textcolor[HTML]{0E9888}{z_{ijb_1b_2}} \in \left\{0, 1\right\}\quad \textnormal{for } \left(i, j\right) \in N, \enskip b_1 \in B_i, \enskip b_2 \in B_j  \label{ilp:binaryz}
\end{eqnarray}}{
  \begin{subequations}
  \newcommand{\setmuskip}[2]{#1=#2\relax}
  \setmuskip{\medmuskip}{0.35mu}
  \setmuskip{\thickmuskip}{0.35mu}
  \setlength{\jot}{1pt}
  \label{equation:ilp}
  \begin{flalign}
    \min \phantom{.} & \textnormal{distance} &\\[-4pt]
    \textnormal{s.t.} \phantom{.} & \textnormal{distance} = \sum_{i=1}^{k} \sum_{b\in{B_i}} d_{ib} \textcolor[HTML]{FA8231}{v_{ib}} & \label{ilp:distance} \\[-4pt]
  & \textcolor[HTML]{E03177}{-S_x} \leq \sum_{i=1}^{k} \sum_{b\in{B_i}} g_{ib} \textcolor[HTML]{FA8231}{v_{ib}} + \sum_{\left(i, j\right) \in N} \sum_{b_1 \in B_i} \sum_{b_2 \in B_j} h_{ijb_1b_2} \textcolor[HTML]{0E9888}{z_{ijb_1b_2}} & \label{ilp:cf} \\[0pt]
  \normalsize
  & \textcolor[HTML]{0E9888}{z_{ijb_1b_2}} = \textcolor[HTML]{FA8231}{v_{ib_1} v_{jb_2}} \hspace{5px} \textnormal{for } \left(i, j\right) \in N, \enskip b_1 \in B_i, \enskip b_2 \in B_j & \label{ilp:interaction}\\[0pt]
  & \sum_{b\in{B_i}}^{} \textcolor[HTML]{FA8231}{v_{ib}} \leq 1 \hspace{22px} \textnormal{for } i = 1, \dots, k & \label{ilp:one}\\[0pt]
  & \textcolor[HTML]{FA8231}{v_{ib}} \in \left\{0, 1\right\} \hspace{25px} \textnormal{for } i = 1, \dots, k, \enskip b \in B_i  & \label{ilp:binaryv}\\[0pt]
  & \textcolor[HTML]{0E9888}{z_{ijb_1b_2}} \in \left\{0, 1\right\} \hspace{12px} \textnormal{for } \left(i, j\right) \in N, \enskip b_1 \in B_i, \enskip b_2 \in B_j & \label{ilp:binaryz}
  \end{flalign}
\end{subequations}}
\renewcommand\theequation{\arabic{equation}}
\setcounter{equation}{2}
\noindent We use an \tcolor{myorange}{indicator variable $\textcolor[HTML]{FA8231}{v_{ib}}$}~(\ref{ilp:binaryv}) to denote if a main effect bin is active:
if $\textcolor[HTML]{FA8231}{v_{ib}}=1$, we change the \tcolor{myorange}{feature value of $\textcolor[HTML]{FA8231}{x_i}$} to the closest value in its bin $b$.
All bin options of $\textcolor[HTML]{FA8231}{x_i}$ are included in a set $B_i$.
For each \tcolor{myorange}{feature $\textcolor[HTML]{FA8231}{x_i}$}, there can be at most one active bin~(\ref{ilp:one}); if there is no active bin, then we do not change the \tcolor{myorange}{value of $x_i$}.
We use an \tcolor{mygreen}{indicator variable $\textcolor[HTML]{0E9888}{z_{ijb_1b_2}}$}~(\ref{ilp:binaryz}) to denote if a \tcolor{mygreen}{pairwise interaction effect} is active---it is active if and only if bin $b_1$ of \tcolor{myorange}{$\textcolor[HTML]{FA8231}{x_i}$} and bin $b_2$ of \tcolor{myorange}{$\textcolor[HTML]{FA8231}{x_j}$} are both active~(\ref{ilp:interaction}).
The set $N$ includes all available \tcolor{mygreen}{interaction effect terms}.
Constraint~\ref{ilp:distance} determines the total distance cost for a potential CF example; it uses a set of pre-computed distance costs $d_{ib}$ of changing one \tcolor{myorange}{feature $x_i$} to the closest value in bin $b$.
Constraint~\ref{ilp:cf} ensures that any solution would flip the model prediction, by gaining enough total score from main effect scores~($g_{ib}$) and interaction effect scores~($h_{ijb_1b_2}$).
Constants $g_{ib}$ and $h_{ijb_1b_2}$ are pre-computed and adjusted for cases where a single active main effect bin results in changes in interaction terms (see~\autoref{sec:milp:ip} for details).

\mypar{Novelty.}
Advancing existing works that use integer linear programs for CF generation (on linear models~\cite{ustunActionableRecourseLinear2019} or using a linear approximation of neural networks~\cite{mohammadiScalingGuaranteesNearest2021}), our algorithm is the first that works on non-linear models without approximation.
Our algorithm is also the first and only CF method specifically designed for EBM models.
Without it, users would have to rely on model-agnostic techniques such as genetic algorithm~\cite{schleichGeCoQualityCounterfactual2021} and KD-tree~\cite{vanlooverenInterpretableCounterfactualExplanations2020} to generate CF examples.
These model-agnostic methods do not allow for customization.
Also, by quantitatively comparing our method with these two model-agnostic CF techniques on three datasets, we find CFs generated by our method are significantly \textit{closer} to the original input, \textit{more sparse}, and encounter \textit{less failures} (see~\autoref{sec:practical:compare} and \autoref{tab:comparison} for details).

\aptLtoX[graphic=no,type=html]{}{
  \phantomsection
}
\mypar{Generalizability.}\label{sec:method:ip:generalizability}
Our algorithm can easily be adapted for EBM regressors and multiclass classifiers.
For regression, we modify the left side and the inequality of constraint~\ref{ilp:cf} to bound the prediction value in the desired range~(see~\autoref{sec:practical:regression} for details).
For multiclass classification, we can modify constraint~\ref{ilp:cf} to ensure that the desired class has the largest score (see~\autoref{sec:practical:multiclass} for details).
In addition to EBMs, one can also adapt our algorithm to generate CF examples for linear models~\cite{ustunActionableRecourseLinear2019}.
For other non-linear models (e.g., neural networks), one can first use a linear approximation~\cite{mohammadiScalingGuaranteesNearest2021} and then apply our algorithm, verifying suggested recourse plans with respect to the original model.
If the suggested recourse plan would not change the output of the original model, an alternative can be generated by solving the program again with the previous solution blocked.

\mypar{Scalability.}
Modern linear solvers can efficiently solve our integer linear programs.
The complexity of solving an integer linear program increases along two factors: the number of variables and the number of constraints.
In \aptLtoX[graphic=no,type=html]{Equation \ref{equation:ilp}}{\autoref{equation:ilp}}, all variables are binary---making the program easier to solve than a program with non-binary integer variables.
For any dataset, there are always exactly 3 constraints from \ref{ilp:distance}, \ref{ilp:cf}, and \ref{ilp:one}.
The number of constraints from \ref{ilp:interaction} increases along the number of interaction terms $|N|$ and the number of bins per feature $|B_i|$ on these interaction terms.
In practice, $|N|$ and $|B_i|$ are often bounded to ensure EBM are interpretable.
For example, by default the popular EBM library InterpretML~\cite{noriInterpretMLUnifiedFramework2019} bounds $|N| \leq 10$ and $|B_i| \leq 32$.
Therefore, in the worst-case scenario with 10 continuous-continuous interaction terms, there will be at most $10 \times 32 \times 32 = 10,240$ constraints from \ref{ilp:interaction}.
For instance, on the Communities and Crime dataset~\cite{redmondDatadrivenSoftwareTool2002} with 119 continuous features, 1 categorical feature, and 10 pairwise interaction terms, there are about 7.2k constraints and 3.6k variables in our program.
It only takes about 0.5--3.0 seconds to generate a recourse plan using Firefox Browser on a MacBook~(see~\autoref{sec:practical:speed} for details).\looseness=-1

\subsection{Recourse Customization}
\label{sec:method:customization}

With integer linear programming, we can generate recourse plans that reflect a wide range of user preferences~(\aptLtoX[graphic=no,type=html]{\textbf{G2}}{\ref{item:g2}}).
For example, to prioritize a feature that is \textit{easier for a user to change}, we can lower the distance cost $d_{ib}$ for that feature~(\autoref{sec:practical:distance}).
To enforce recourse plans to only change a feature in a user specified \textit{acceptable range}, we can remove out-of-range binary variables $\textcolor[HTML]{FA8231}{v_{ib}}$.
If a user requires the recourse plans to only change \textit{at most $p$ features}, we can add an additional linear constraint $\sum_{i=1}^{k}\sum_{b\in{B_i}} \textcolor[HTML]{FA8231}{v_{ib}} \leq p$.
Finally, with modern linear solvers, we can efficiently generate diverse recourse plans~(\aptLtoX[graphic=no,type=html]{\textbf{G1}}{\ref{item:g1}}) by solving the program multiple times while blocking previous solutions~(see \autoref{sec:practical:regression}\figpart{--}\autoref{sec:practical:constraint} for details).

\section{User Interface}
\label{sec:ui}

\begin{figure*}[tb]
  \includegraphics[width=\textwidth]{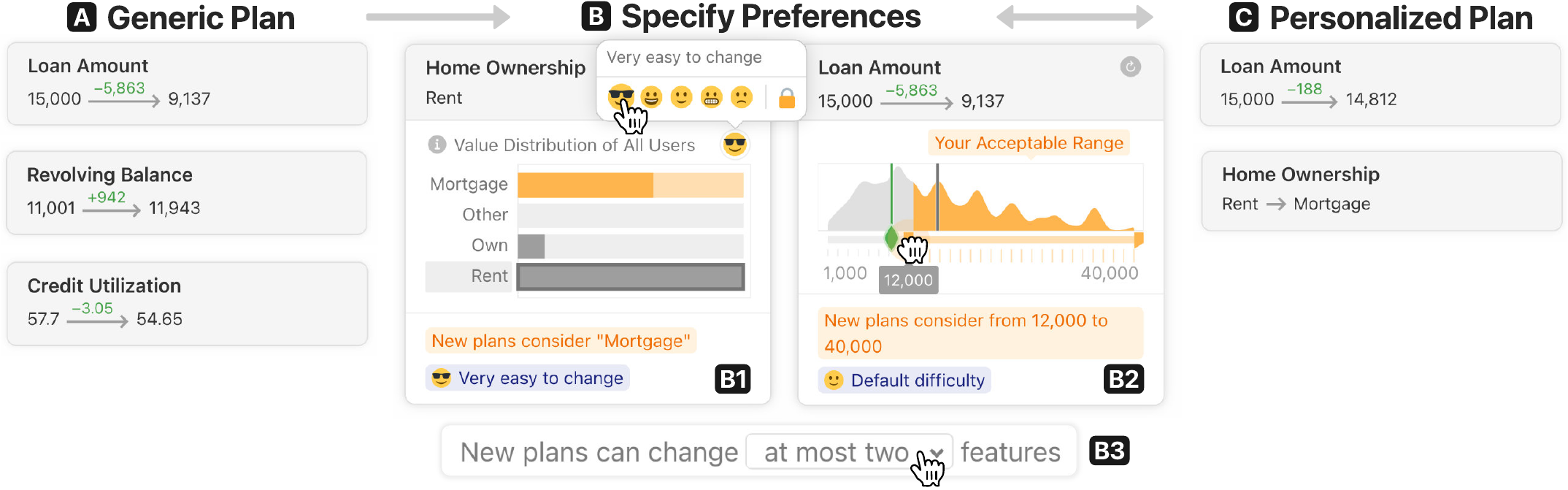}
  \caption[]{
    \tool{} enables end users to inspect and customize recourse plans through simple interactions.
    \textbf{(A)} \textit{Initial generic plans} are generated with the same configurations for all users.
    \textbf{(B)} Users can \textit{specify recourse preferences} if they are not satisfied with the initial plans; by configuring \textbf{(B1)} the \textit{difficulty} to change a feature; \textbf{(B2)} the \textit{acceptable range} that a feature can change between, and \textbf{(B3)} the \textit{max number of features} that a recourse plan can alter.
    \textbf{(C)} \tool{} then generates \textit{personalized plans} that respect users' preferences.
    Users can iteratively refine their preferences until a satisfactory plan is found.
  }
  \Description{
    A flow chart of three screenshots labeled A, B, and C.
    Screenshot A shows three collapsed feature cards under the title ``Generic Plan''.
    Screenshot B shows two extended feature cards under the title ``Specify Preferences''.
    The left is a categorical feature card when a user is specifying the difficulty.
    The right is a continuous feature card when a user is changing its acceptable range.
    Screenshot C shows two collapsed feature cards under the title ``Personalized Plan''.
    Screenshot A points to screenshot B.
    Screenshots B and C point to each other.
  }
  \label{fig:scenario}
\end{figure*}

Given the design goals~(\aptLtoX[graphic=no,type=html]{\textbf{G1}--\textbf{G5}}{\ref{item:g1}--\ref{item:g5}}) described in \autoref{sec:goal}, we present \tool{}, an interactive tool that empowers end users to specify preferences and iteratively fine-tune recourse plans~(\autoref{fig:scenario}).
The interface tightly integrates three components: the \menu{} that provides overall controls and organizes multiple recourse plans as tabs~(\autoref{sec:ui:tab}), the \panel{} containing \cards{} that allow users to specify recourse preferences with simple interactions~(\autoref{sec:ui:panel}), and the \textit{Bookmark Window} summarizing saved recourse plans~(\autoref{sec:ui:bookmark}).
To explain these views in this section, we use a loan application scenario with the LendingClub dataset~\cite{LendingClubOnline2018}, where a bank refers a rejected loan applicant to \tool{} pre-loaded with the applicant's input data.
Our tool can be easily applied to GAMs trained on different datasets while providing a consistent user experience.
On \tool{}'s \linkhref{https://poloclub.github.io/gam-coach/}{public demo page}, we present five additional examples with five datasets that are commonly used in algorithmic recourse literature:
Communities and Crime~\cite{redmondDatadrivenSoftwareTool2002} (also used in the second usage scenario in \autoref{sec:ui:scenario}),
Taiwan Credit~\cite{yehComparisonsDataMining2009}, German Credit~\cite{duaUCIMachineLearning2017}, Adult~\cite{kohaviScalingAccuracyNaivebayes1996}, and COMPAS~\cite{larsonHowWeAnalyzed2016}.
\looseness=-1

\subsection{Coach Menu}
\label{sec:ui:tab}

\begin{figure}[b]
  \includegraphics[width=0.92\linewidth]{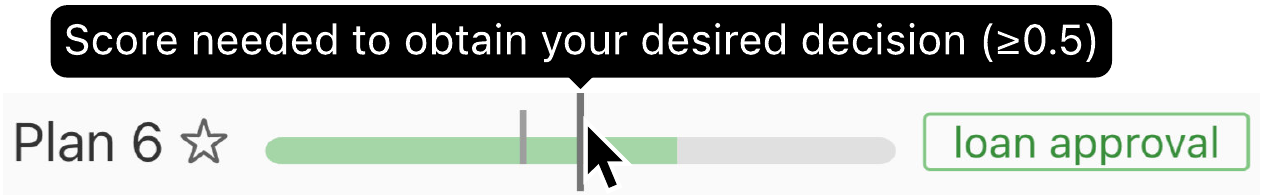}
  \caption[]{
    A bar chart visualizes model's decision score of a recourse plan: the bar is marked with the user's original score (shorter vertical line on the left) and the threshold needed to obtain the desired decision (longer vertical line on the right).
  }
  \Description{
    A screenshot of a small bar chart in the plan tab bar.
    There is a green bar between the text labels ``Plan 6'' and ``loan approval''.
    A cursor is on top of the midpoint of the green bar with a tooltip ``Score needed to obtain your desired decision (>= 0.5).''
  }
  \label{fig:score-tab}
\end{figure}

\setlength{\belowcaptionskip}{5pt}
\setlength{\abovecaptionskip}{12pt}
\setlength{\belowcaptionskip}{0pt}
\setlength{\abovecaptionskip}{12pt}

The \menu{}~(\autoref{fig:teaser}\figpart{A}) is the primary control panel of \tool{}.
Users can use the dropdown menu and input fields to specify desired decisions for classification and regression.
For each recourse plan generation iteration, the tool generates five diverse plans~(\aptLtoX[graphic=no,type=html]{\textbf{G1}}{\ref{item:g1}}) to help users achieve their goal, with each plan representing a CF example.
Users can access each plan by clicking the corresponding tab on the plan tab bar.
When a plan is selected, the \panel{} updates to show details about the plan, and the plan's corresponding tab extends to show the model's decision score~(\autoref{fig:score-tab}).
Users can click the \vcenteredhbox{\includegraphics[height=10pt]{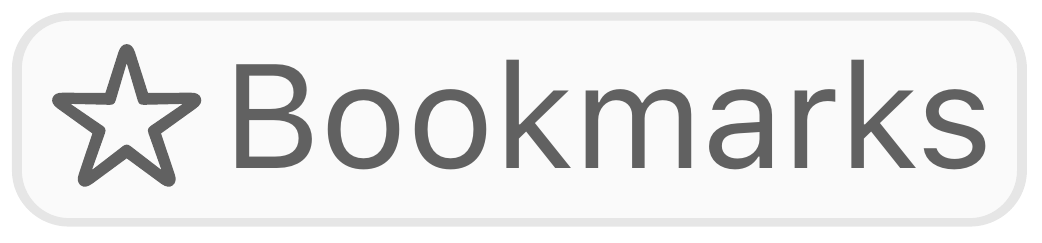}} button to open the \mybookmark{} window and click the \vcenteredhbox{\includegraphics[height=10pt]{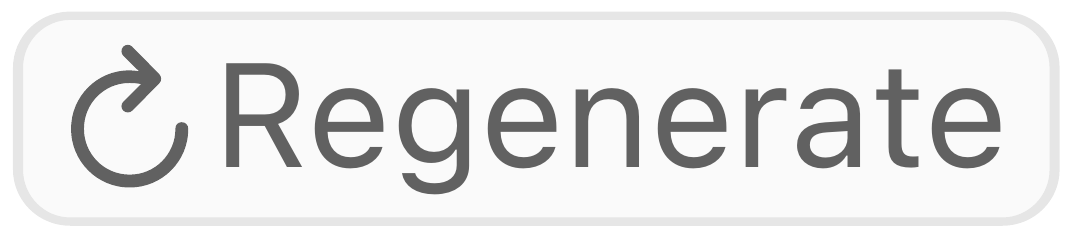}} button to generate five new recourse plans that reflect the currently specified recourse preferences.

\subsection{Feature Panel}
\label{sec:ui:panel}

Each recourse plan has a unique \panel{}~(\autoref{fig:teaser}\figpart{B}) that visualizes plan details and allows users to provide preferences guiding the generation of new plans~(\aptLtoX[graphic=no,type=html]{\textbf{G2}}{\ref{item:g2}}).
A \panel{} consists of \cards{} where each card represents a data feature used in the model.
To help users easily navigate through different features, the panel groups \cards{} into three sections: (1) features that are changed in the plan, (2) features that are configured by the user, (3) and all other features.
To prevent overwhelming users with too much information~(\aptLtoX[graphic=no,type=html]{\textbf{G4}}{\ref{item:g4}}), all cards are collapsed by default---only displaying the feature name and feature values.
Users can hover over the feature name to see a tooltip explaining the definition of the feature~(\aptLtoX[graphic=no,type=html]{\textbf{G4}}{\ref{item:g4}}).
With a \textit{progressive disclosure} design~\cite{shneidermanEyesHaveIt1996, normanUserCenteredSystem1986}, details of a feature, such as the distribution of feature values, are only shown on demand after users click that \card{}.
Progressive disclosure also makes \tool{} interface scalable, as users can easily scroll and browse over hundreds of collapsed \cards{}.
Since EBMs process continuous and categorical features differently, we employ different card designs based on the feature type.

\setlength{\columnsep}{10pt}%
\setlength{\intextsep}{-2pt}%
\begin{wrapfigure}{R}{0.27\textwidth}
  \vspace{0pt}
  \centering
  \includegraphics[width=0.27\textwidth]{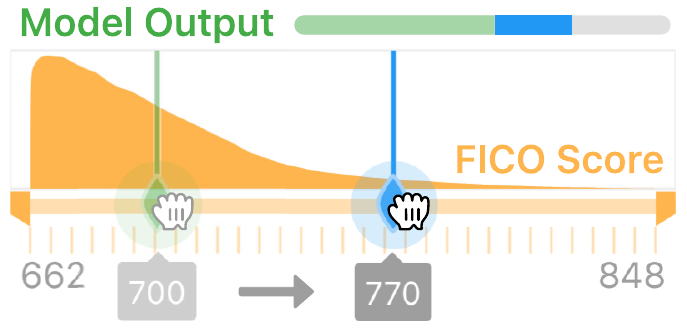}
  \vspace{-25pt}
  \caption[]{
    Users can test hypothetical input values in real time.
  }
  \label{fig:whatif}
  \Description{
    A screenshot of the filled curve plot of FICO Score.
    A user is changing the current feature value from 800 to 770
    The model output bar chart's width extends.
  }
\end{wrapfigure}
\mypar{Continuous Feature Card.}
For continuous features, such as \vcenteredhbox{\includegraphics[height=9pt]{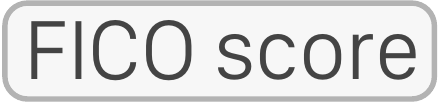}}, the \card{}~(\autoref{fig:whatif}) uses a filled curved chart to visualize the distribution of feature values in the training set.
Users can drag the diamond-shaped thumb \vcenteredhbox{\includegraphics[height=10pt]{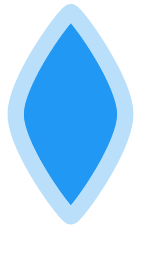}} on a slider below the chart to experiment with hypothetical values.
During dragging, the decision score bar updates its width to reflect a new prediction score in real time.
Therefore, users can better understand the underlying decision-making process by probing the model with different inputs~(\aptLtoX[graphic=no,type=html]{\textbf{G3}}{\ref{item:g3}}).
Also, users can drag the orange thumbs \vcenteredhbox{\includegraphics[height=7pt]{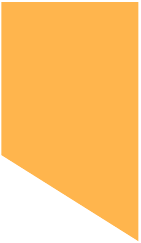}} \vcenteredhbox{\includegraphics[height=7pt]{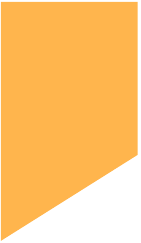}} to set the lower and upper bounds of acceptable feature changes.
For example, one user might only accept recourse plans that include \vcenteredhbox{\includegraphics[height=9pt]{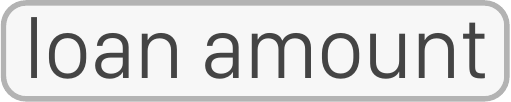}} at \$12k or higher~(\autoref{fig:scenario}\figpart{-B2}).

\mypar{Categorical Feature Card.}
For categorical features, such as \vcenteredhbox{\includegraphics[height=9pt]{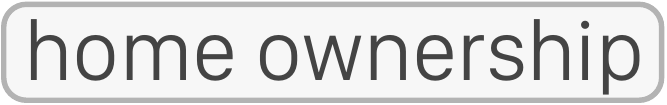}}, users can inspect the value distribution with a horizontal bar chart~(\autoref{fig:scenario}\figpart{-B1}), where a longer bar represents more frequent options in the training data.
To specify acceptable ranges, users can click the bars to select or deselect acceptable options for new recourse plans.
Acceptable options are highlighted as \orangehl{orange}, whereas unacceptable options are colored as \grayhl{gray}.
Users can also click text labels next to the bars to experiment with hypothetical options and observe how they affect the model decision.

\setlength{\columnsep}{11pt}%
\setlength{\intextsep}{0pt}%
\begin{wrapfigure}{R}{0.18\textwidth}
  \vspace{0pt}
  \centering
  \includegraphics[width=0.175\textwidth]{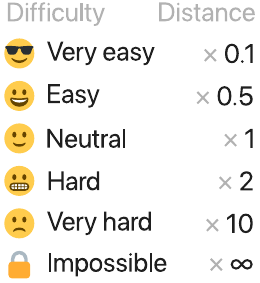}
  \vspace{-8pt}
  \caption[]{
    Distance multipliers of difficulties.
  }
  \Description{
    A table mapping difficulty icons to their distance multipliers.
    Very easy maps to ``⨉ 0.1''.
    Easy maps to ``⨉ 0.5''.
    Neutral maps to ``⨉ 1''.
    Hard maps to ``⨉ 2''.
    Very hard maps to ``⨉ 10''.
    Impossible maps to ``⨉ infinity''.
  }
  \label{fig:diff-map}
\end{wrapfigure}
\mypar{Specify Difficulty to Change a Feature.}
Besides selecting a feature's acceptable range, users can also specify how hard it would be for them to change a feature.
For example, it might be easier for some users to lower \vcenteredhbox{\includegraphics[height=9pt]{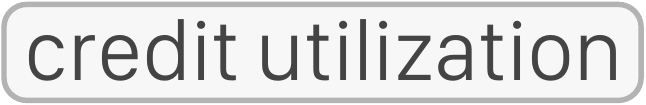}} than to change \vcenteredhbox{\includegraphics[height=9pt]{figures/feature-home}}.
To configure feature difficulties, users can click the smiley button on any \card{} and then select a suitable difficulty option on the pop-up window~(\autoref{fig:scenario}\figpart{-B1}).
Internally, \tool{} multiplies the distance costs of all bins in that feature with a constant multiplier~(\autoref{fig:diff-map}).
If the user selects the ``impossible to change'' difficulty, the tool will remove all variables associated with this feature in the internal integer program~(\autoref{sec:method:customization}).
Therefore, when generating new recourse strategies, \tool{} would prioritize features that are easier to change and would not consider features that are impossible to change.\looseness=-1

\setlength{\belowcaptionskip}{0pt}
\setlength{\abovecaptionskip}{5pt}
\begin{figure*}[tb]
  \includegraphics[width=\textwidth]{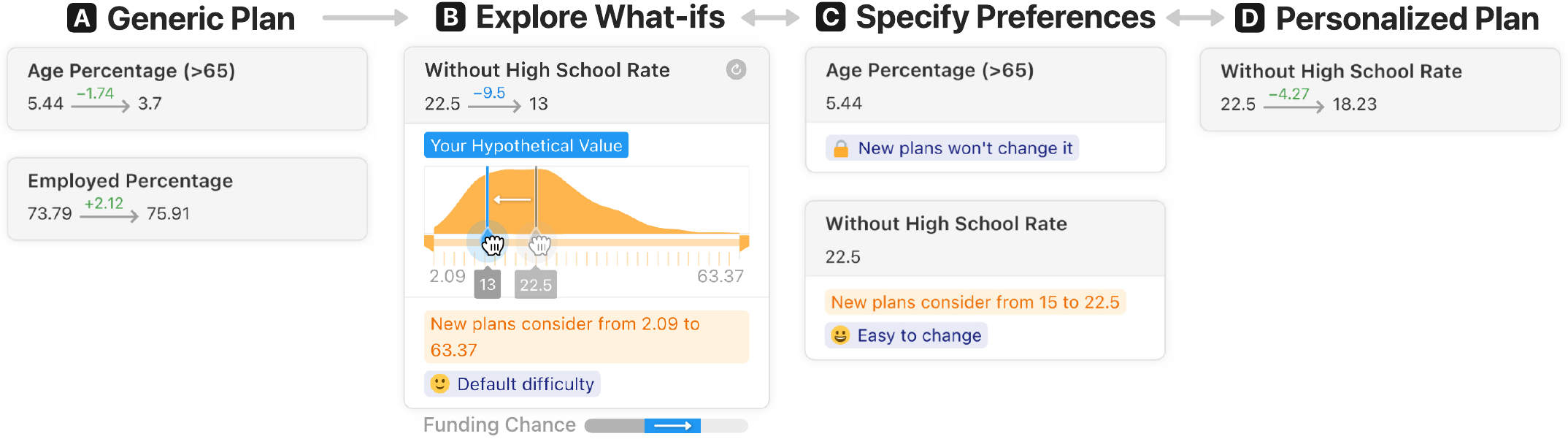}
  \caption[]{
    \tool{} allows end users to experiment with hypothetical input values and customize recourse plans.
    \textbf{(A)} Our tool first shows \textit{generic plans} generated with default configurations.
    \textbf{(B)} Users can explore how different input values affect the model's prediction in real time through simple interactions on the \card{}: for example, lowering the percentage of adults without a high school diploma increases the chance of getting a government grant.
    \textbf{(C)} Users can then specify recourse preferences---such as feature \textit{difficulties} and \textit{acceptable ranges}---based on their circumstances and understanding of the model's prediction patterns.
    \textbf{(D)} \tool{} then generates more actionable recourse plans based on the user-specified preferences.
  }
  \Description{
    A flow chart of three screenshots labeled A, B, C, and D.
    Screenshot A shows two collapsed feature cards (``age percentage (>65)'' and ``employed percentage'') under the title ``Generic Plan''.
    Screenshot B shows an extended feature card (``without high school rate'') under the title ``Explore what-ifs''.
    The user is dragging the slider to explore a hypothetical value on this feature.
    Screenshot C shows two collapsed feature cards under the title ``Specify Preferences''.
    The user sets ``age percentage (>65)'' to ``new plans won't change it'' and ``without high school rate'' to ``easy to change''.
    Screenshot D shows a collapsed feature of ``without high school rate'' under the title ``Personalized Plan''.
    Screenshot A points to screenshot B.
    Screenshots B and C point to each other.
    Screenshots C and D point to each other.
  }
  \label{fig:scenario-crime}
\end{figure*}
\setlength{\belowcaptionskip}{0pt}
\setlength{\abovecaptionskip}{12pt}

\subsection{Bookmarks and Receipt}
\label{sec:ui:bookmark}

During the recourse iterations, users can save any suitable plans by clicking the star button \vcenteredhbox{\includegraphics[height=8pt]{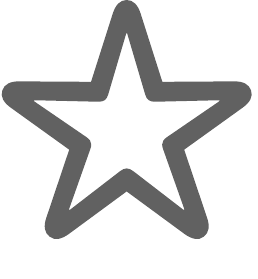}} on the plan tab~(\autoref{fig:score-tab}).
Then, users can compare and update their saved plans in the \mybookmark{} \textit{window}~(\autoref{fig:teaser}\figpart{C}).
Once users are satisfied with bookmarked plans, they can save a \textit{recourse receipt} as proof of the generated recourse plans.
\citet{wachterCounterfactualExplanationsOpening2017} first introduced the recourse receipt concept as a contract guaranteeing that a bank will approve a loan application if the applicant achieves all changes listed in the recourse plan.
\tool{} is the first tool to realize this concept by creating a plaintext file that records the timestamp, a hash of EBM model weights, the user's original input, and details of bookmarked plans~(\aptLtoX[graphic=no,type=html]{\textbf{G4}}{\ref{item:g4}}).
In addition, we propose a novel security scheme that uses Pretty Good Privacy (PGP) to sign the receipt with the bank's private key~\cite{garfinkelPGPPrettyGood1995}.
With public-key cryptography, users can hold the bank accountable by being able to prove the receipt's authenticity to third-party authorities with the bank's public key.
Also, banks can use their private key to verify a receipt's integrity during recourse redemption to avoid counterfeit receipts.\looseness=-1

\subsection{Usage Scenarios}
\label{sec:ui:scenario}

We present two hypothetical usage scenarios to illustrate how \tool{} can potentially help everyday users identify actionable strategies to alter undesirable ML-generated decisions.

\mypar{Individual Loan Application.}
Eve is a rejected loan applicant, and she wants to identify ways to get a loan in the future.
In this hypothetical usage scenario, to inform loan decisions, the bank has trained an EBM model on past data (we use LendingClub~\cite{LendingClubOnline2018} to illustrate this scenario in \autoref{fig:scenario}).
Their dataset has 9 continuous features and 11 categorical features~(\autoref{fig:appendix-input}), and the outcome variable is binary---indicating whether a person can pay back the loan in time.
The bank gives Eve a \linkhref{https://poloclub.github.io/gam-coach/?dataset=lending}{link to \tool{}} when informing her of the loan rejection decision.
After Eve opens \tool{} in a web browser, the tool pre-loads Eve's input data and generates five recourse plans based on the default configurations.
Each plan lists a set of minimal changes in feature values that would lead to loan approval.
One plan suggests Eve lower the requested \vcenteredhbox{\includegraphics[height=9pt]{figures/feature-amount}} from \$15k to \$9k along with two other changes~(\autoref{fig:scenario}\figpart{A}).
Eve does not like this suggestion because she is unwilling to compromise a loss of \$6k in the requested loan.
Therefore, she clicks the \vcenteredhbox{\includegraphics[height=9pt]{figures/feature-amount}} \card{} and drags the left thumb~\vcenteredhbox{\includegraphics[height=7pt]{figures/icon-thumb1}} to set the \textit{acceptable range} of \vcenteredhbox{\includegraphics[height=9pt]{figures/feature-amount}} to \$12k and above~(\autoref{fig:scenario}\figpart{-B2}).
After browsing all recourse plans in the \menu{}, Eve finds that none of the plans suggest changes to \vcenteredhbox{\includegraphics[height=9pt]{figures/feature-home}}.
Eve and her partner are actually moving to their newly-purchased condo next month.
Therefore, Eve sets the \textit{acceptable range} of \vcenteredhbox{\includegraphics[height=9pt]{figures/feature-home}} to ``mortgage'' and changes its \textit{difficulty} to ``very easy'' ~\vcenteredhbox{\includegraphics[height=10pt]{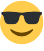}}~(\autoref{fig:scenario}\figpart{-B1}).
Eve also prefers plans that change fewer features, so she clicks the dropdown menu on the \panel{} to ask the tool to only generate plans that change at most two features~(\autoref{fig:scenario}\figpart{-B3}).
After Eve clicks the \vcenteredhbox{\includegraphics[height=10pt]{figures/icon-regenerate}} button, \tool{} quickly generates five personalized plans that respect Eve's preferences.
Among these plans, Eve especially likes the one suggesting she lower the \vcenteredhbox{\includegraphics[height=9pt]{figures/feature-amount}} by about \$200 and change \vcenteredhbox{\includegraphics[height=9pt]{figures/feature-home}} to mortgage~(\autoref{fig:scenario}\figpart{C}).
Finally, Eve bookmarks this plan and downloads a recourse receipt that guarantees her a loan if all suggested terms are met.
Eve plans to apply for the loan again at the same bank next month.

\mypar{Government Grant Application.}
Hal is a county manager in the United States.
He has applied for a federal grant for his county.
Unfortunately, his application is rejected.
He wants to learn about the decision-making process and what actions he can take to succeed in future applications.
In this hypothetical usage scenario, to inform funding decisions, the federal government has trained an EBM model on past data (we use the Communities and Crime dataset~\cite{redmondDatadrivenSoftwareTool2002} to illustrate this scenario in \autoref{fig:scenario-crime}).
This dataset has 119 continuous features and 1 categorical feature describing the demographic and economic information of different counties in the United States, and is used to predict the risk of violent crime.
As part of a performance incentive funding program~\cite{verainstituteofjusticePerformanceIncentiveFunding2012},
the federal government provides more funding opportunities to counties with lower predicted crime risk~\cite{slackCounterfactualExplanationsCan2021}.
Before training the EBM model, the federal government has removed protected features (e.g., \vcenteredhbox{\includegraphics[height=9pt]{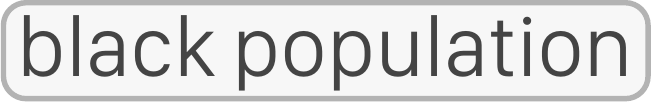}}) and features with many (more than half) missing values, resulting in a total of 94 continuous features and 1 categorical feature.

The federal government provides rejected counties with a \linkhref{https://poloclub.github.io/gam-coach/?dataset=crime}{link to \tool{}} when informing them of the funding decisions.
Hal opens \tool{} in his browser; this tool has pre-loaded the demographic and economic features of his county and quickly suggested five recourse plans that would lead to funding.
These generic plans are generated with the default configuration.
One plan~(\autoref{fig:scenario-crime}\figpart{A}) suggests Hal decrease \vcenteredhbox{\includegraphics[height=9pt]{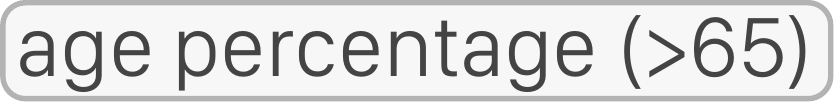}} and increase \vcenteredhbox{\includegraphics[height=9pt]{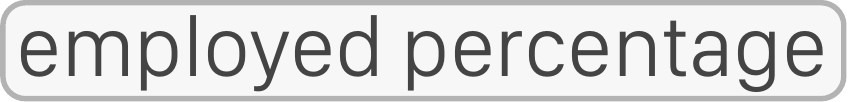}} in his county.
Hal likes the recommendation of increasing \vcenteredhbox{\includegraphics[height=9pt]{figures/feature-employed}} because a higher employment rate is also beneficial for the economy of his county.
However, Hal is puzzled by the suggestion of lowering \vcenteredhbox{\includegraphics[height=9pt]{figures/feature-age-65}}.
He is not sure why the population age is used to decide funding decisions.
Besides, lowering the percentage of the elderly population is not actionable.
Therefore, Hal ``locks'' this feature by setting its \textit{difficulty} to ``impossible''~\vcenteredhbox{\includegraphics[height=9pt]{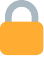}}~(\autoref{fig:scenario-crime}\figpart{C}).

To gain a better understanding of how the funding decision is made, Hal expands several \cards{} and experiments with hypothetical feature values by dragging the blue thumbs \vcenteredhbox{\includegraphics[height=10pt]{figures/icon-thumb2}}; \tool{} visualizes the model's prediction scores with these hypothetical inputs in real time~(\autoref{fig:scenario-crime}\figpart{B}).
Hal quickly finds that lowering \vcenteredhbox{\includegraphics[height=9pt]{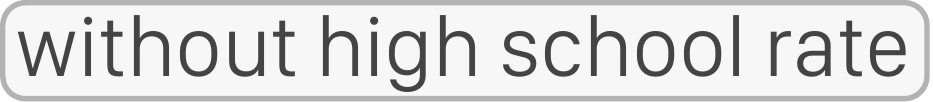}} can increase his chance of getting a grant.
This is good news as Hal's county has just started a high school dropout prevention program aiming to lower the percentage of adults without a high school diploma to below 15\% in eight years.
Hal then sets this feature's \textit{difficulty} to ``easy to change''~\vcenteredhbox{\includegraphics[height=9pt]{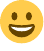}} and drags the orange thumbs~\vcenteredhbox{\includegraphics[height=7pt]{figures/icon-thumb1}} \vcenteredhbox{\includegraphics[height=7pt]{figures/icon-thumb3}} to set its \textit{acceptable range} to between 15\% and 22.5\%~(\autoref{fig:scenario-crime}\figpart{C}).
After Hal clicks the \vcenteredhbox{\includegraphics[height=10pt]{figures/icon-regenerate}} button, \tool{} generates five new personalized plans in only 3 seconds despite there being almost 100 features.
Among these five plans, Hal likes the one that recommends decreasing \vcenteredhbox{\includegraphics[height=9pt]{figures/feature-highschool}} by 4.27\%~(\autoref{fig:scenario-crime}\figpart{D}).
Finally, Hal saves a recourse receipt, and he will apply for this grant again once the percentage of adults without a high school diploma in his county drops by 4.27\%.

\subsection{Open-source \& Generalizable Tool}
\label{sec:ui:implement}

\tool{} is a web-based algorithmic recourse tool that users can access with any web browser on their laptops or tablets, no installation required~(\aptLtoX[graphic=no,type=html]{\textbf{G5}}{\ref{item:g5}}).
We use \textit{GLPK.js}~\cite{vaillantGlpkJs2021} to solve integer programs with WebAssembly, \textit{OpenPGP.js}~\cite{haseOpenPGPJsOpenPGP2014} to sign recourse receipts with PGP, and \textit{D3.js}~\cite{bostockDataDrivenDocuments2011} for visualizations.
Therefore, the entire system runs locally in users' browsers without dedicated backend servers.
We also provide an additional Python package\footnote{Python package: \link{https://poloclub.github.io/gam-coach/docs/gamcoach}} for developers to generate customizable recourse plans for EBM models without a graphical user interface.
With this Python package, developers and researchers can also easily extract model weights from any EBM model to build their own \tool{}.
Finally, despite its name, \tool{}'s interface is model-agnostic---it supports any ML models where (1) one can control the difficulty and acceptable range of changing a feature during CF generation, and (2) model inference is available.
With our open-source and generalizable implementation, detailed documentation, and examples on six datasets across a wide range of tasks and domains---LendingClub~\cite{LendingClubOnline2018}, Taiwan Credit~\cite{yehComparisonsDataMining2009}, German Credit~\cite{duaUCIMachineLearning2017}, Adult~\cite{kohaviScalingAccuracyNaivebayes1996}, COMPAS~\cite{larsonHowWeAnalyzed2016}, and Communities and Crime~\cite{yehComparisonsDataMining2009}---future researchers can easily adapt our interface design to their models and datasets. %
\section{User Study}
\label{sec:user}

To evaluate \tool{} and investigate how everyday users would use an interactive algorithmic recourse tool, we conducted an online user study with 41 United States-based crowdworkers.
For possible datasets to use in this user study, we compared five public datasets that are commonly used in the recourse literature:
LendingClub~\cite[e.g.,][]{mothilalExplainingMachineLearning2020,tsirtsisDecisionsCounterfactualExplanations2020}, Taiwan Credit~\cite[e.g.,][]{tsirtsisDecisionsCounterfactualExplanations2020,ustunActionableRecourseLinear2019,schleichGeCoQualityCounterfactual2021}, German Credit~\cite[e.g.,][]{mothilalExplainingMachineLearning2020,tsirtsisDecisionsCounterfactualExplanations2020,slackCounterfactualExplanationsCan2021}, Adult~\cite[e.g.,][]{karimiModelAgnosticCounterfactualExplanations2020,schleichGeCoQualityCounterfactual2021,mohammadiScalingGuaranteesNearest2021}, and COMPAS~\cite[e.g.,][]{mothilalExplainingMachineLearning2020,karimiModelAgnosticCounterfactualExplanations2020,rawalIndividualizedRecourseInterpretable2020}.
We decided to use LendingClub in our study for the following three reasons.
First, we chose a lending scenario as it is one scenario that many people, including crowdworkers, may encounter in real-life.
Second, there is no expert knowledge needed to understand the setting, making our tasks appropriate for crowdworkers.
Finally, our institute requires research participants to be United States-based: among the four datasets that can be used in a lending setting (LendingClub, Taiwan Credit, German Credit, and Adult), LendingClub is the only United States-based dataset collected from a real lending website.
In this user study, we aimed to answer the following three research questions:

\aptLtoX[graphic=no,type=html]{
\begin{itemize}
  \item[\textbf{RQ1.}] What makes a satisfactory recourse plan for end users? (\autoref{sec:user:result1})

  \item[\textbf{RQ2.}] How do end users discover their satisfactory recourse plans? (\autoref{sec:user:result2})

  \item[\textbf{RQ3.}] How does interactivity play a role in providing algorithmic recourse? (\autoref{sec:user:result3})
\end{itemize}
}{
\begin{enumerate}[topsep=5pt, itemsep=0mm, parsep=1mm, leftmargin=24pt, label=\textbf{RQ\arabic*.}, ref=RQ\arabic*]
  \item \label{item:q1} What makes a satisfactory recourse plan for end users? (\autoref{sec:user:result1})

  \item \label{item:q2} How do end users discover their satisfactory recourse plans? (\autoref{sec:user:result2})

  \item \label{item:q3} How does interactivity play a role in providing algorithmic recourse? (\autoref{sec:user:result3})
\end{enumerate}
}

\subsection{Participants}

We recruited 50 anonymous and voluntary United States-based participants from Amazon Mechanical Turk (MTurk), an online crowdsourcing platform.
We did not collect any personal information. Collected interaction logs and subjective ratings are stored in a secure location where only the authors have access.
The authors' Institutional Review Board (IRB) has approved the study.
The average of three self-reported task completion times on a worker-centered forum\footnote{TurkerView: \link{https://turkerview.com/}} is 32\nicefrac{1}{2}-minutes.
We paid 41 participants \$6.50 per study and 9 participants who had not passed our quality control \$5.50.\footnote{Originally the task was posted with a base payment of \$3.50 and \$1 bonus for quality. However, when analyzing participants' responses, we realized that the task required more time than we originally expected, so we provided an additional \$2 bonus to all participants after the study to ensure appropriate compensation for their time. This brought the payment to \$6.50 for those who passed the quality control quiz and \$5.50 for those who did not.}
Recruited participants self-report an average score of 2.7 for ML familiarity in a 5-point Likert-scale, where 1 represents ``I have never heard of ML'' and 5 represents ``I have developed ML models.''

\subsection{Study Design}

To start, each participant signed a consent form and filled out a background questionnaire (e.g., familarity with ML).

\setlength{\belowcaptionskip}{0pt}
\setlength{\abovecaptionskip}{5pt}
\begin{figure}[tb]
  \includegraphics[width=0.8\linewidth]{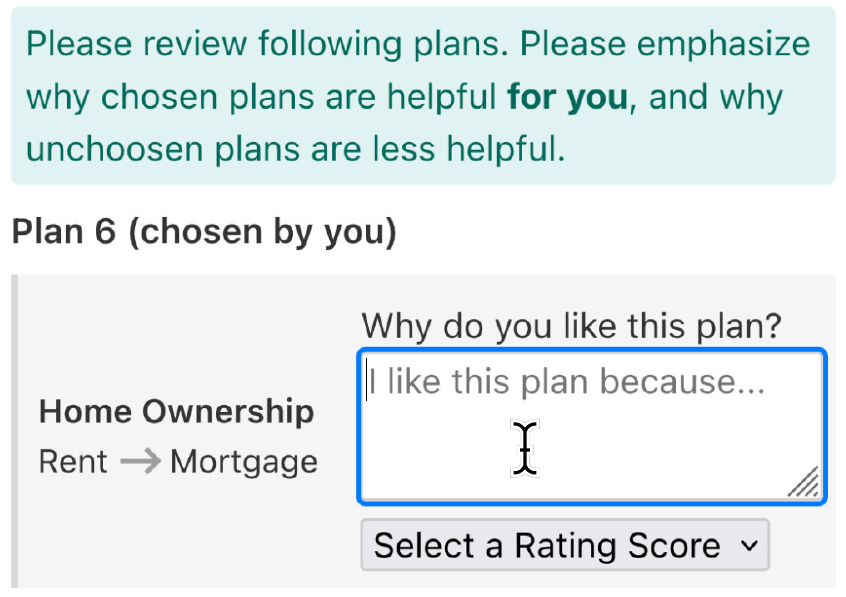}
  \caption[]{We asked user study participants to explain why they had chosen their satisfactory plans, and why they had not chosen two other random plans (not shown in figure).}
  \Description{
    A screenshot of a user study question.
    The instruction writes ``Please review the following plans. Please emphasize why chosen plans are helpful for you, and why unchosen plans are less helpful.''
    There is a box below the instruction labeled ``Plan 6 (chosen by you).''
    Below the label, there is a recourse plan with the feature Home Ownership changed from Rent to Mortgage.
    On the right of the plan, there is a question ``Why do you like this plan?''
    There is a text field and rating drop-down menu below the question.
  }
  \label{fig:plan-review}
\end{figure}
\setlength{\belowcaptionskip}{0pt}
\setlength{\abovecaptionskip}{12pt}

\mypar{\tool{} Tutorial and Short Quiz.}
We directed participants to a Google Survey form and \linkhref{https://poloclub.github.io/gam-coach/user-study/}{a website} containing \tool{}, task instructions, and tutorial videos.
Our tool, loaded with an EBM binary classifier that predicts loan approval on the LendingClub dataset~\cite{LendingClubOnline2018}, also contains input values of 500 random test samples on which the model predicts loan rejection.
Participants were asked to watch a 3-minute tutorial video and complete eight multiple-choice quiz questions.
These questions are simple---asking what is shown in the tool after certain interactions.
All participants were asked to perform these interactions on the same data sample, so we had ``ground truth'' answers for the quiz questions.
We used the quiz as a ``gold standard'' question to detect fraudulent responses~\cite{olsonWaysKnowingHCI2014, kitturCrowdsourcingUserStudies2008}.
Although participants were prompted that they would need to answer all questions correctly to receive the base compensation, we paid all participants regardless of their answers.
However, in our analysis, we only included responses from participants who had correctly answered at least four questions.

\mypar{Free Exploration with an Imaginary Usage Scenario.}
After completing the tutorial and quiz, participants were asked to pretend to be a rejected loan applicant and freely use \tool{} \textit{until finding at least one satisfactory recourse plan.}
These satisfactory recourse plans could be chosen from the first five generic plans that \tool{} generates with a default configuration \textit{or} follow-up plans that are generated based on participants' configured preferences.
To help participants imagine the scenario, we asked them to change the input sample (one of 500 random samples) until they find one that they feel comfortable pretending to be.
Participants could also manually adjust the input values~(\autoref{fig:appendix-input} in the appendix).
After identifying and bookmarking their satisfactory plans, participants were asked to rate the importance of configured preferences or briefly explain why no configuration is needed.
Then, participants were asked to explain why they had chosen their saved plans~(\autoref{fig:plan-review}) and why they had not chosen two other plans, which were randomly picked from the initial recourse plans.
To incentivize participants to write good-quality explanations~\cite{paolacciTurkUnderstandingMechanical2014, hoIncentivizingHighQuality2015}, we told participants that they could get a \$1 bonus reward if their explanations are well-justified.
Regardless of their responses, all participants who had correctly answered at least four quiz questions were rewarded with this bonus.\looseness=-1

\mypar{Interaction Logging and Survey.}
While participants were using \tool{}, the tool logged all interactions, such as preference configuration, hypothetical value experiment, and recourse plan generation.
Each log event includes a timestamp and associated values.
After finishing the exploration task, participants were asked to click a button that uploads their interaction logs and recourse plan reviews as a JSON file to a secured Dropbox directory.
The filenames included a random number.
Participants were given this number as a verification code to report in the survey response and MTurk submission---we used this number to link a participant's MTurk ID with their log data and survey response.
Finally, participants were asked to complete the survey consisting of subjective ratings and open-ended comments regarding the tool.
As the EBM model used in the study is non-monotonic,
the tool sometimes can suggest counterintuitive changes~\cite{barocasHiddenAssumptionsCounterfactual2020}, such as to lower \vcenteredhbox{\includegraphics[height=9pt]{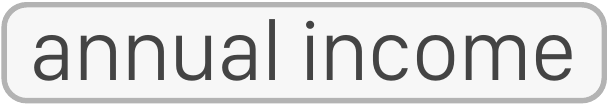}} for loan approval.
We asked participants to report counterintuitive recourse plans in the survey if they had seen any.

\subsection{Results}

Out of 50 recruited participants, 41 (P1--P41) correctly answered at least four ``quality-control'' questions.
In the following sections, we summarize our findings through analyzing these 41 participants' interaction logs, recourse plan reviews, and survey responses.
We denote the Wald Chi-Square statistical test score as $\chi^2$.

\subsubsection{RQ1: Characteristics of Satisfactory Recourse Plans}
\label{sec:user:result1}

During the exploration task, participants were asked to identify at least one recourse plan that they would be satisfied with if they were a rejected loan applicant using \tool{}.
On average, each participant chose 1.54~\vcenteredhbox{\includegraphics[height=9pt]{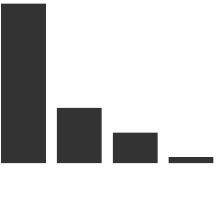}} satisfactory plans.
Participants preferred \textit{concise plans} that changed only a few features, with an average of 2.11~~\vcenteredhbox{\includegraphics[height=9pt]{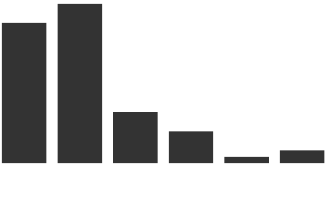}} features per plan.
Chosen plans changed a diverse set of features, including 13 out of 20 features.
The most popular features changed by chosen plans were \vcenteredhbox{\includegraphics[height=9pt]{figures/feature-amount}}~(26.3\%), \vcenteredhbox{\includegraphics[height=9pt]{figures/feature-fico}}~(18.8\%), and \vcenteredhbox{\includegraphics[height=9pt]{figures/feature-utilization}}~(11.3\%).
Features that were not changed by any chosen plans were mostly hard to change in real life, such as \vcenteredhbox{\includegraphics[height=9pt]{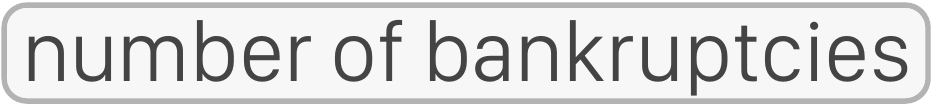}} and \vcenteredhbox{\includegraphics[height=9pt]{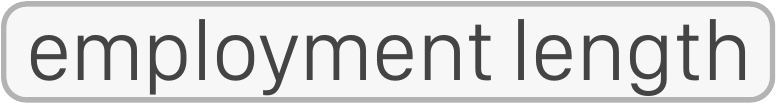}}.

\aptLtoX[graphic=no, type=html]{}{
  \setul{0.4ex}{0.3ex}
  \setulcolor{mygray}
}

\mypar{Reasons for Choosing Satisfactory Plans.}
Three main reasons that participants reported choosing plans were that the plans were (1) controllable, (2) requiring small changes or less compromise, or (3) beneficial for life in general.
Most participants chose recourse plans that felt realistic and controllable.
For example, P30 wrote \myquote{I think it's very possible to reduce my \ul{credit utilization} in a short amount of time.}
In particular, participants preferred plans that only changed a few features and required a small amount of change.
Participants described these plans as ``\textit{simple and fast}'' (P5), ``\textit{straightforward}'' (P7), and ``\textit{easy to do}'' (P16).
Some participants chose plans because they could tolerate the compromises.
For example, P8 wrote \myquote{I'm fine with the lower \ul{loan amount}.}
Similarly, P11 reported \myquote{[The decreased] \ul{loan amount} is close to what I need.}
Interestingly, some participants favored plans that could benefit their lives in addition to helping them get loan approval.
For example, P14 wrote \myquote{[...] lower \ul{utilization} is good for me anyway from what I know, so this seems like the best plan.}
Similarly, P28 wrote \myquote{[this plan] in my opinion would guarantee greater monetary flexibility.} \looseness=-1

\mypar{Reasons for Not Choosing a Plan.}
Participants' explanations for not choosing a plan mostly complemented the reasons for choosing a plan.
Some participants also skipped plans because they were puzzled by counterintuitive suggestions, did not understand the suggestions, or just wanted to see more alternatives.
First, participants disliked unrealistic suggestions:
P2 explained \myquote{It tells me to increase my \ul{income}. My \ul{income} is fixed. I cannot just increase them at a whim.}
Similarly, P6 wrote \myquote{With inflation it might be harder to \ul{use less credit}.}
Participants also disliked plans requiring too many changes or a large amount of change.
For example, P30 wrote \myquote{The \ul{amount of loan} suggested to be reduced is too large. Assuming I'm applying for 9,800 for real, I wouldn't want to reduce the amount by more than 30\%.}
Interestingly, some participants skipped a plan because it suggested counterintuitive changes.
For example, P14 wrote \myquote{It seemed like a bug because why would asking for an extra 13 dollars [in \ul{loan amount}] result in a loan approval?}
Participants also skipped plans when they did not understand the suggestion: P9 wrote \myquote{I'm not exactly sure what \ul{credit utilization} is. I looked at the tooltip, but still wasn't sure.}
Finally, some participants skipped the initial plans because they just wanted to explore more alternatives: P22 explained \myquote{I wanted to check out a few more things before I made my decision.}\looseness=-1

\mypar{Design Lessons.}
By analyzing the characteristics of satisfactory recourse plans, our user study is the first study that provides empirical evidence to support several hypotheses from the recourse literature.
We find that participants preferred plans that suggested changes on actionable features~\cite{karimiSurveyAlgorithmicRecourse2021,kirfelWhatIfHow2021}, are concise and make small changes~\cite{leGRACEGeneratingConcise2020,wachterCounterfactualExplanationsOpening2017}, and could benefit participants beyond the recourse goal~\cite{barocasHiddenAssumptionsCounterfactual2020}.
Additionally, participants were likely to save multiple satisfactory plans from one recourse session, highlighting the importance of providing diverse recourse plans~\cite{mothilalExplainingMachineLearning2020}.
Our study also shows that with transparency, end users can identify and dislike counterintuitive recourse plans (see more discussion in \autoref{sec:user:result3}).
Therefore, future researchers and developers should help users identify concise and diverse plans that change actionable features and are beneficial overall.
Also, researchers and developers should carefully audit and improve their models to prevent a CF generation algorithm from generating counterintuitive plans.
Our findings also highlight that communicating recourse plans and providing a good user experience are as important as generating good recourse plans.

\subsubsection{RQ2: Path to Discover Satisfactory Recourse Plans}
\label{sec:user:result2}

In the exploration task, participants could freely choose their satisfactory recourse plans from the initial batch, where plans were generated with default configurations, or from follow-up batches, where plans reflected participants' specified preferences.
We find that participants were more likely to choose satisfactory plans that respect participants' preference configurations (33 participants out of 41) than the default plans (8 participants).
In addition, each recourse session had a median of 3~\vcenteredhbox{\includegraphics[height=9pt]{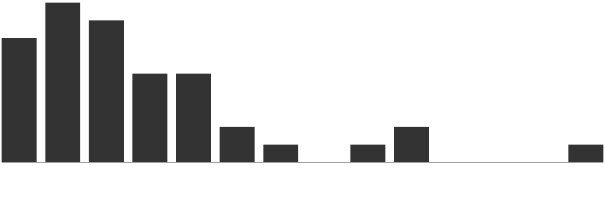}} plan iterations.
In other words, on average, a participant discovered satisfactory plans after seeing about 15 plans, where the last 10 plans were generated based on their preferences.
The average time to identify satisfactory plans was 8 minutes and 38 seconds.

\mypar{Preference configuration is helpful.}
In \tool{}, users can specify the \textit{difficulty} and \textit{acceptable range} to change a feature and the \textit{max number of features} a plan can change.
We find all three preferences helped participants discover satisfactory plans.
Among 63 total satisfactory plans chosen by 41 participants, 49 plans~(77.78\%) reflected at least one difficulty configuration and 44 plans~(69.84\%)  reflected at least one range configuration.
Also, 12 participants configured the max number of features---seven participants changed it to 1 and five changed it to 2 (default is 4).

\setlength{\belowcaptionskip}{0pt}
\setlength{\abovecaptionskip}{5pt}
\begin{figure}[tb]
  \includegraphics[width=0.95\linewidth]{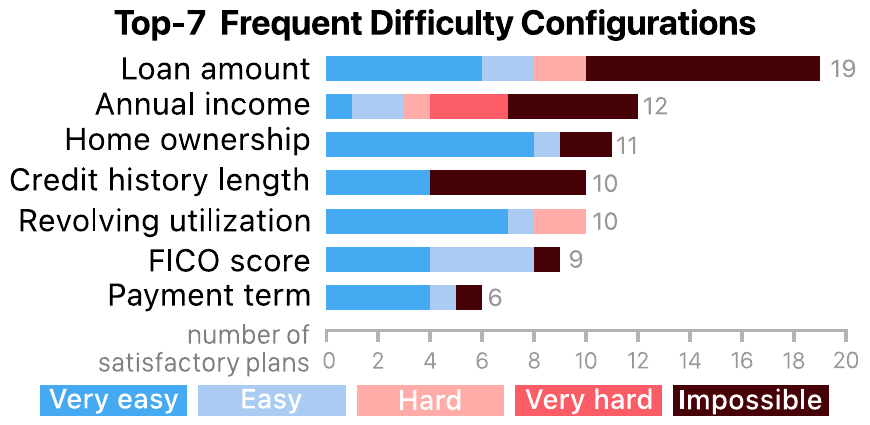}
  \caption[]{
    Difficulty configuration counts across frequent features highlighting
    variability of participants' preferences.
  }
  \Description{
    A horizontal stacked bar chart of the top 7 frequent difficulty configurations.
    From the top to the bottom, the feature loan amount has been used in 19 satisfactory plans,
    annual income in 12 plans, home ownership in 11 plans, credit history length in 10 plans,
    revolving utilization in 10 plans, FICO score in 9 plans, and payment term in 6 plans.
  }
  \label{fig:result-preference}
\end{figure}
\setlength{\belowcaptionskip}{0pt}
\setlength{\abovecaptionskip}{12pt}

\mypar{Diverse Preference Configurations.}
By further analyzing participants' preferences associated with their chosen plans, we find (1) participants specified preferences on a wide range of features; (2) some features were more popular than others; (3) different participants set different preferences on a given feature.
Of the 20 features, at least one participant changed the difficulty of 16 features (80\%) and acceptable range of 13 features (65\%).
Among these configured features, participants were more likely to specify preferences on some than others [$\chi^2 = 54.37$, $p<0.001$ for the difficulty, $\chi^2 = 27.68$, $p=0.006$ for the acceptable range].
For example, 19 satisfactory plans reflected difficulty for \vcenteredhbox{\includegraphics[height=9pt]{figures/feature-amount}}, whereas only 1 plan reflected the difficulty for \vcenteredhbox{\includegraphics[height=9pt]{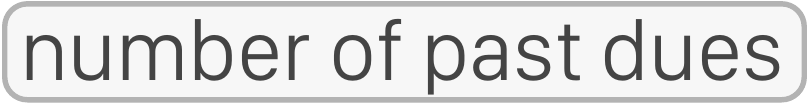}}.
Also, there was high variability in configured preferences on popular configured feature~(\autoref{fig:result-preference}).
For instance, 6 plans considered \vcenteredhbox{\includegraphics[height=9pt]{figures/feature-amount}} as ``very easy to change,'' while 9 plans deemed it as ``impossible to change.''
Our findings confirm hypotheses that recourse preferences can be incorporated to identify satisfactory plans~\cite{barocasHiddenAssumptionsCounterfactual2020, weldChallengeCraftingIntelligible2019}, and these preferences are idiosyncratic~\cite{kirfelWhatIfHow2021,vermaCounterfactualExplanationsMachine2020}.

\mypar{Design Lessons.}
When designing recourse systems, it is useful to allow end users to specify a wide range of recourse preferences, such as difficulties to change a feature, acceptable feature ranges, and max number of features to change.
Additionally, there can be predictable patterns in users' recourse preferences---researchers can leverage these patterns to further improve user experiences.
For example, developers can use the log data of an interactive recourse tool to train a new ML model to predict users' preference configurations.
Then, for a new user, developers can predict their recourse preference and use it as the tool's default configuration.

\subsubsection{RQ3: Interactive Algorithmic Recourse}
\label{sec:user:result3}

\begin{figure*}[tb]
  \includegraphics[width=\textwidth]{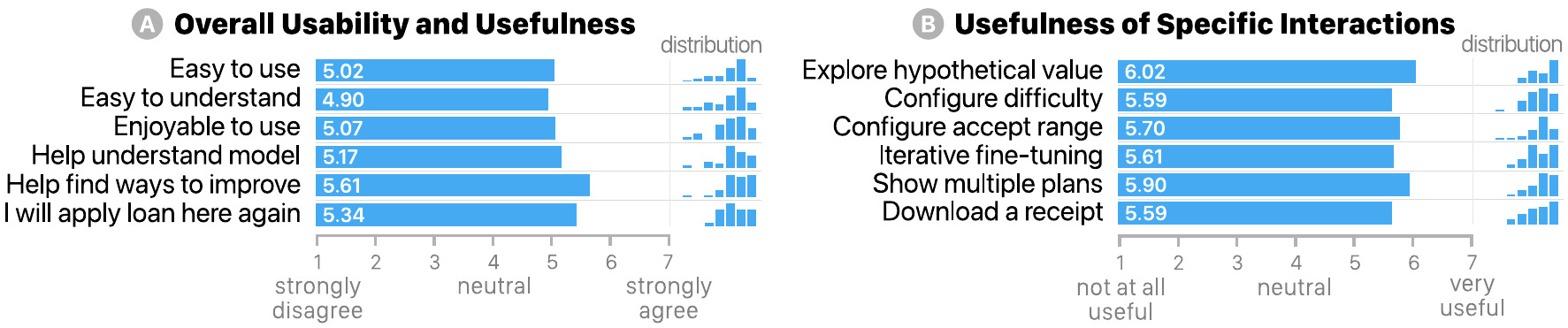}
  \caption[]{
    Average ratings and rating distributions from 41 participants on the usability and usefulness of \tool{}.
    \textbf{(A)} Participants thought \tool{} was relatively easy and enjoyable to use, and the tool helped them identify actions to obtain a preferred ML decision.
    \textbf{(B)} All interaction techniques, especially experimenting with hypothetical values, were rated favorably.
  }
  \Description{
    Two horizontal bar charts of usability and usefulness average ratings.
    For usability, ``easy to use'' has an average score of 5.02, ``easy to understand'' of 4.9,
    ``enjoyable to use'' of 5.07, ``help understand model'' of 5.17, ``help find ways to improve''
    of 5.61, and ``I will apply loan here again'' of 5.34,
    For usefulness, ``explore hypothetical value'' has an average score of 6.02, ``configure difficulty''
    of 5.59, ``configure accept range'' of 5.7, ``iterative fine-tuning'' of 5.61, ``show multiple plans'' of 5.9, and ``download a receipt'' of 5.59.
  }
  \label{fig:usability}
\end{figure*}

How did participants use and perceive various \textit{interactions} throughout the exploration task?
Interestingly, 28\% of participants who configured difficulty preferences had also immediately altered the difficulty levels on the same features; most of them have changed ``easy'' to ``very easy'' and ``hard'' to ``very hard.''
For acceptable ranges, the percentage is higher at 88\%.
It suggests participants may need iterations to learn how preference configuration works in \tool{} and then fine-tune configurations to generate better plans---highlighting the key role of iteration in interactive recourse.
Survey response show that participants found both preference configuration and iteration helpful in finding good recourse plans~(\autoref{fig:usability}\figpart{B}).
For example, P30 commented \myquote{[I like] how easy it was to make changes to the priority of each thing. Showing that some things can be easy changes, or impossible to change, and making plans built around those.}
Similarly, P19 wrote \myquote{[I like] regenerating unlimited plans until I find a fit one.}

\setlength{\belowcaptionskip}{0pt}
\setlength{\abovecaptionskip}{8pt}
\setlength{\belowcaptionskip}{0pt}
\setlength{\abovecaptionskip}{12pt}

\mypar{``What-if'' Questions.}
Besides configuring preferences, participants also engaged in other modes of interaction with \tool{}.
For example, 32 out of 41 participants experimented with hypothetical feature values~(\autoref{sec:ui:panel}), even though it did not affect recourse generations and was not required in the task.
These participants explored median of 3 unique features~\vcenteredhbox{\includegraphics[height=9pt]{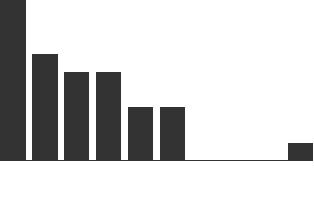}} and a median of 5.5 hypothetical feature values~\vcenteredhbox{\includegraphics[height=9pt]{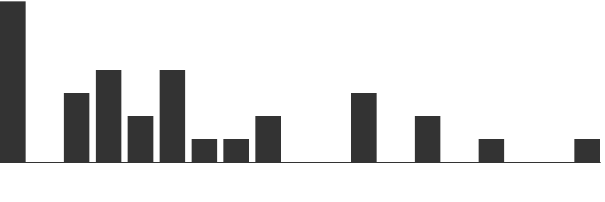}}.
These 32 participants asked what-if questions on a total of 99 features, and only 39~(39.4\%) of these features were from the presented recourse plan.
It suggests that participants were more interested in learning about the predictive effects of features that have not been changed by \tool{}.
After exploring what-ifs on these 99 features, participants configured at least one preference (difficulty or acceptable range) on about half of them~(49 features, 49.5\%).
In comparison, these participants only configured preferences on 13.72\% features~(87 out of 634) on which they had not explored what-ifs or had explored what-ifs \textit{after} configuring preferences.
It shows that participants were more likely to customize features on which they had explored hypothetical values [$\chi^2 = 85.459$, $p<0.00001$].
Finally, 20 out of these 32 participants~(62.5\%) chose a satisfactory plan with a changed feature on which they had explored what-ifs.
It may suggest participants preferred recourse plans that changed features on which they had explored what-ifs, but this result is not statistically significant [$\chi^2 = 2.0$, $p=0.1573$].

By analyzing survey responses, we also find that asking what-if questions was one of the participants' favorite features~(\autoref{fig:usability}\figpart{B}).
For example, P12 wrote \myquote{[I like] how it adjusts the plans in real time and gives you an answer if the loan will be approved.}
Throughout the task, participants also frequently used the tooltip annotations to inspect the decision score bar (median 8 times per participant) and check the meaning of different features (median 25 times)---highlighting the importance of clearly explaining visual representations and terminologies in interactive recourse tools. \looseness=-1

\mypar{Counterintuitive recourse plans.}
We asked participants to report strange recourse plans that \tool{} could rarely suggest, such as to lower \vcenteredhbox{\includegraphics[height=9pt]{figures/feature-income}} for loan approval.
To our surprise, 7 out of 41 participants had encountered and reported these counterintuitive plans!
For example, P6 was confused that some plans suggested conflicting changes on the same feature: \myquote{One plan told me to increase the \ul{loan amount} by \$13 while another plan told me to decrease by \$1,613.}
Another interesting case was P39: \myquote{I don't understand how \ul{purpose} changes approval decision. Something like \ul{`mortgage'} I understand, but changing something and all of a sudden you can do a wedding but not home improvement? Like what?}
First, P39 found it counterintuitive that \tool{} includes the categorical feature \vcenteredhbox{\includegraphics[height=9pt]{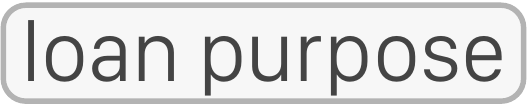}} as a changeable feature because they thought the model decision should be independent of the \vcenteredhbox{\includegraphics[height=9pt]{figures/feature-purpose}}.
Then, through experimenting with hypothetical values, P39 was baffled by the observation that two different purposes (wedding and home improvement) resulted in two distinct model decisions.
Some other participants also attributed these strange patterns as reasons why they skipped some plans~(\autoref{sec:user:result1}).
This finding provides empirical evidence that with transparency, everyday users can discover potentially problematic behaviors in ML models.

\mypar{Design Lessons.}
Overall, interactivity helps users identify satisfactory recourse plans, and users appreciate being able to control recourse generation.
In addition, users like being able to ask what-if questions; experimenting with hypothetical feature values also helps them find satisfactory recourse plans.
However, it takes time and trial and error for users to understand how preference configurations affect recourse generation.
Therefore, future interactive recourse tools can improve user experience by focusing on improving learnability and reversibility.
Also, our study shows that interactivity and transparency could occasionally confuse users with counterintuitive recourse plans.
Therefore, future researchers and developers should carefully audit and improve their ML models before deploying interactive recourse tools.

\subsubsection{Usability}
\label{sec:user:usability}

Our survey included a series of 7-point Likert-scale questions regarding the usability of \tool{}~(\autoref{fig:usability}\figpart{A}).
The results suggest that the tool is relatively easy to use~(average 5.02), easy to understand~(average 4.90), and enjoyable to use~(average 5.07).
However, some participants commented that the tool was not easy to learn at first and may be too complex for users with less knowledge about loans.
For example, P5 wrote \myquote{Without the tutorials, it would have taken me much longer to learn how to navigate the program, because it is not very intuitive at first.}
Similarly, P8 wrote \myquote{I am decent with finances, but I'd imagine that other people would have more difficulty [using the tool].}
Our participants were MTurk workers, who are similar to the demographics of American internet users as a whole, but slightly younger and more educated~\cite{olsonWaysKnowingHCI2014, hitlinResearchCrowdsourcingAge2016}.
Therefore, \tool{} might be overwhelming for real-life loan applicants who are less familiar with web technology or finance.
Participants also provided specific feedback for improvement, such as designing a better way to \textit{store} and \textit{compare} all generated plans.
Currently, users would lose unsaved plans when generating new plans, and users could only compare different recourse plans in the \mybookmark{} \textit{window}~(\autoref{sec:ui:bookmark}).
We plan to continue improving the design of \tool{} based on participants' feedback.
\section{Limitations}

We acknowledge limitations regarding our tool's generalizability, usage scenarios, and user study design.

\mypar{Generalizability of \tool{}.}
To design and develop the first interactive algorithmic recourse tool that enables end users to fine-tune recourse plans with preferences, we ground our research in GAMs, a class of accurate and transparent ML models with simple structures.
This approach enables us to generate customizable CF examples efficiently.
However, not all CF generation algorithms allow users to specify the feature-level distance functions, acceptable ranges, and max number of features that a CF example can change.
Therefore, while the \tool{} interface is model-agnostic, it does not directly support all existing ML models and CF generation methods.
Also, our novel CF generation algorithm is tailored to EBMs.
However, one can easily adapt our linear constraints to generate customizable CF examples for linear models~\cite{ustunActionableRecourseLinear2019}.
For more complex non-linear models (e.g., random forest, neural networks), one can apply our method to a linear approximation~\cite{mohammadiScalingGuaranteesNearest2021} of these models~(\autoref{sec:method:ip:generalizability}).
We also acknowledge that similar to most existing CF generation algorithms~\cite{keaneIfOnlyWe2021,barocasHiddenAssumptionsCounterfactual2020}, our algorithm assumes all features to be independent.
However, in practice, many features can be associated.
For example, changing \vcenteredhbox{\includegraphics[height=9pt]{figures/feature-utilization}} is likely to also affect a user's \vcenteredhbox{\includegraphics[height=9pt]{figures/feature-fico}}.
Future work can generalize our algorithm to dependent features by modeling their casual relationships~\cite{karimiAlgorithmicRecourseCounterfactual2021}.

\mypar{Hypothetical Usage Scenarios.}
We situate \tool{} in lending and government funding settings~(\autoref{sec:ui:scenario}), two most cited scenarios in existing CF literature~\cite{karimiSurveyAlgorithmicRecourse2021,barocasHiddenAssumptionsCounterfactual2020}.
It is important to note that none of the authors have expertise in law, finance, or political science.
Therefore, to adapt \tool{} for use in real lending and government funding settings, it would require more research and engaging with experts in the legal and financial domains as well as people who would be impacted by the systems.
In addition, we use LendingClub~\cite{LendingClubOnline2018} and Communities and Crime~\cite{redmondDatadrivenSoftwareTool2002}, two largest suitable datasets we have access to~(\autoref{sec:user}), to simulate two usage scenarios and design our user study.
These two datasets can have different features and sizes from the data that are used in practice.
Therefore, before adapting \tool{}, researchers and developers should thoroughly test our tool on their own datasets.

\mypar{Simulated Study Design.}
To study how end users would use interactive recourse tools, we recruited MTurk workers and asked them to pretend to be rejected loan applicants, and we logged and analyzed their interactions with \tool{}.
We designed the task to encourage and help participants simulate the scenario (e.g., rewarding bonus, supporting participants to input data or choose data from multiple random samples).
However, participants' usage patterns and reactions may not fully represent real-life loan applicants.
We chose to simulate a lending scenario because (1) crowdworkers may have encountered lending, (2) it does not require expert knowledge, and (3) we have access to a large and real US-based lending dataset.
We acknowledge that participants' usage patterns may not full represent users in other domains.
Therefore, it would require further research with actual end users (e.g., loan applicants, county executives, and bail applicants) to study how \tool{} can aid them in real-world settings.
In our study, we only collected participants' familiarity with ML.
As MTurk workers tend to be younger and more educated than average internet users~\cite{olsonWaysKnowingHCI2014, hitlinResearchCrowdsourcingAge2016}, future researchers can collect more self-reported demographic information (e.g., age, education, sex) to study if different user groups would use an interactive recourse tool differently.

\mypar{Observational Study Design.}
Our observational log study can provide a portrait of users' natural behaviors when interacting with interactive algorithmic recourse tools and scale to a large number of participants~\cite{dumaisUnderstandingUserBehavior2014}.
However, it lacks a control group.
As algorithmic recourse research and applications are still nascent, the community has not yet established a recommended workflow or system that we can use as a baseline in our study~(\autoref{sec:related:recourse}).
Our main goal is to study how \textit{recourse customizability} can help users discover useful recourse plans.
Therefore, to mitigate the lack of a control group, we offer participants the option to \textit{abstain from customizing recourse plans} to probe into the usefulness of recourse customizability.
In our analysis, we compare both (1) the numbers of participants who specify recourse preferences and who do not, (2) and the numbers of satisfactory plans generated with a default configuration and satisfactory plans generated with a participant-configured preference~(\autoref{sec:user:result2}).
Finally, with our open-source implementation~(\autoref{sec:ui:implement}), future researchers can use \tool{} as a baseline system to evaluate their interactive recourse tools.

\section{Discussion and Future Work}

Reflecting on our end-to-end realization of interactive algorithmic recourse---from UI design to algorithm development and a user study---we distill lessons and provide a set of future directions for algorithmic recourse and ML interpretability.

\mypar{Too much transparency.}
\tool{} uses a glass-box model, provides end users with complete control of recourse plan generation, and supports users to ask ``what-if'' questions with any feature values.
One might argue that \tool{} is too transparent and too much transparency makes the tool unfavorable, because (1) end users can use this tool for gaming the ML model~\cite{kleinbergHowClassifiersInduce2020, hardtStrategicClassification2016} and (2) this tool fails to protect the decision maker's model intellectual property~\cite{wachterCounterfactualExplanationsOpening2017}.
We acknowledge these concerns.
As recourse research and applications are still nascent, it is challenging to know how we can balance the benefits of transparency and human agency and the risk of revealing too much information about the ML model.
Our user study shows that with transparency end users can discover and are often puzzled by counterintuitive patterns in ML models.
We believe if \tool{} is adopted, it has the potential to incentivize decision makers to create better models in order to avoid confusion as well as model exploitations.
As one of the furthest realizations of ML transparency, \tool{} can be a research instrument that facilitates future researchers to study the tension between \textit{decision makers} and \textit{decision subjects}, and identify the right amount of transparency that most benefits both parties.
Then, to adopt \tool{} in practice, ML developers can remove certain functionalities or impose recourse constraints accordingly.
For example, if a bank is offering \tool{} and is worried about people gaming the system by changing certain features that do not actually improve their creditworthiness (e.g., opening more credit cards), they could insert their own optimization constraints that prevent these features from being modified.

\mypar{Transparent ML models for algorithmic recourse.}
Black-box ML models are popular across different domains.
To interpret these models, researchers have developed post-hoc techniques to identify feature importance~\cite[e.g.][]{ribeiroWhyShouldTrust2016,lundbergUnifiedApproachInterpreting2017} and generate CF examples~\cite[e.g.][]{leGRACEGeneratingConcise2020,mothilalExplainingMachineLearning2020}.
However, \citet{rudinStopExplainingBlack2019} argues that researchers and practitioners should use transparent ML models instead of black-box models in high-stake domains due to transparent models' high accuracy and explanation fidelity.
The design of \tool{} is based on GAMs, a state-of-the-art transparent model~\cite{caruanaIntelligibleModelsHealthCare2015,wangPursuitInterpretableFair2020}.
We would like to broaden the perspective of using transparent models reflecting on our study.
We find that \tool{} provides opportunities for everyday users to discover counterintuitive patterns in the ML model.
It implies that ML developers and researchers can also use \tool{} as a penetration testing tool to detect potentially problematic behaviors in their models.
Note that both black-box and transparent learning methods would have learned these counterintuitive behaviors~\cite{caruanaIntelligibleModelsHealthCare2015}, but with a transparent model, developers can further \textit{vet} and \textit{fix} these behaviors.
As an example, an ML developer training a GAM can use \tool{} to iteratively generate recourse plans for potential users (e.g., training data where the model gives unfavorable predictions).
If they identify strange suggestions, they can use existing interactive tools~\cite{noriInterpretMLUnifiedFramework2019,wangInterpretabilityThenWhat2022} to visualize corresponding shape functions to pinpoint the root cause of these counterintuitive patterns, and then edit shape function parameters to avoid them from happening during recourse deployment.
Future research can leverage transparent models to distill guidelines to audit and fix models before recourse deployment.\looseness=-1

\mypar{Put users at the center.}
During the design and implementation of \tool{}, we have encountered many challenges in transforming technically sound recourse plans into a seamless user experience.
As the end users of recourse tools are everyday people who are less familiar with ML and domain-specific concepts, one of our design goals is to help them understand necessary concepts and have a frictionless experience~(\aptLtoX[graphic=no,type=html]{\textbf{G4}}{\ref{item:g4}}).
\tool{} aims to achieve this goal by following a progressive disclosure and details-on-demand design strategy~\cite{normanUserCenteredSystem1986,shneidermanEyesHaveIt1996} and presenting textual annotations to explain visual representations in the tool.
However, our user study suggests that few users might still find it challenging to use \tool{} at first~(\autoref{sec:user:usability}).
During our development process, we identify many edge cases that a recourse application would encounter in practice, such as features requiring integer values (e.g., \vcenteredhbox{\includegraphics[height=9pt]{figures/feature-fico}}), features using log transformations (e.g., \vcenteredhbox{\includegraphics[height=9pt]{figures/feature-income}}), or features less familiar to everyday users (e.g., \vcenteredhbox{\includegraphics[height=9pt]{figures/feature-utilization}}).
Our open-source implementation handles these edge cases, and we provide ML developers with simple APIs to add descriptions for domain-specific feature names in their own instances of \tool{}.
However, these practical edge cases are rarely discussed or handled in the recourse research community, since (1) the field of algorithmic recourse is relatively nascent, (2) and the main evaluation criteria of recourse research are distance-based statistics instead of \textit{user experience}~\cite{keaneIfOnlyWe2021}.
Therefore, in addition to developing faster techniques to generate more actionable recourse plans, we hope future researchers engage with end users and incorporate user experience into their research agenda.
Besides interactive visualization, researchers can also explore alternative mediums to communicate and personalize ML recourse plans and model explanations, such as through a textual~\cite{ehsanRationalizationNeuralMachine2018} or multi-modal approach~\cite{hohmanTeleGamCombiningVisualization2019}.
\section{Conclusion}

As ML models are increasingly used to inform high-stakes decision-making throughout our everyday life, it is crucial to provide decision subjects ways to alter unfavorable model decisions.
In this work, we present \tool{}, an interactive algorithmic recourse tool that empowers end users to specify their preferences and iteratively fine-tune recourse plans.
Our tool runs in web browsers and is open-source, broadening people's access to responsible ML technologies.
We discuss lessons learned from our realization of interactive algorithmic recourse and an online user study.
We hope our work will inspire future research and development of user-centered and interactive tools that help end users restore their human agency and eventually trust and enjoy ML technologies. 

\begin{acks}
We thank Kaan Sancak for his support in piloting our user study.
We appreciate Harsha Nori, Paul Koch, Samuel Jenkins, and the InterpretML team for answering our questions about InterpretML.
We express our gratitude to our study participants for testing our tool and providing valuable feedback.
We are also grateful to our anonymous reviewers for their insightful comments and suggestions that have helped us refine our work.
This work was supported in part by a J.P. Morgan PhD Fellowship, gifts from Bosch and Cisco.
\end{acks}

\balance
\bibliographystyle{ACM-Reference-Format}
\bibliography{coach}
\appendix
\clearpage

\setcounter{figure}{0}
\renewcommand{\thetable}{S\arabic{table}}
\renewcommand{\thefigure}{S\arabic{figure}}

\section{Recourse Generation Details}

\subsection{EBM CF Generation Problem Definition}
\label{sec:milp:prob}

Given a trained EBM model $M$ and an instance $x \in \mathbb{R}^k$, our goal is to generate a set of CF examples $\{c^{\left(1\right)}, c^{\left(2\right)}, \dots, c^{\left(l\right)}\}$, where $M$ gives a different decision than the original input $x$.
In other words, we would like to find $c$ such that $M\left(c\right) \neq M\left(x\right)$.
Without loss of generality, we use binary classification as an example in this section.
For binary classifications, EBM use sigmoid function $\sigma(a) = \frac{1}{1 + e^{-a}}$ as a link function.
This link function rescales the sum of shape function values $S_x = \beta_0 + f_1\left(x_1\right) + f_2\left(x_2\right) + \cdots + f_k\left(x_k\right) + \cdots + f_{i,j}\left(x_i, x_j\right)$ to a probability $\sigma\left(S_x\right)$, ranging from 0 to 1.
If $\sigma\left(S_x\right) \geq 0.5$ or $S_x \geq 0$, $M$ predicts the input $x$ as \texttt{positive}; otherwise $M$ predicts $x$ as \texttt{negative}.
To generate a CF example $c$ that leads to a different decision than the original input $x$, we need to make some changes to $x$ so that the new score $S_c$ has a different sign from $S_x$.

\subsection{Counterfactual Constraint}
\label{sec:milp:cf}

A CF example $c$ is valid if it changes the sign of the original score $S_x$.
If the model predicts the original input $x$ as positive ($s_x \geq 0$), then the score gain $g\left(x, c\right) = S_c - S_x$ should be smaller than $-S_x$.
Similarly, if the model predicts $x$ as negative ($s_x < 0$), then the score gain $g\left(x, c\right)$ should be at least $-S_x$.
Since EBM is additive during inference, we can write $g\left(x, c\right)$ as:

\begin{equation}
  \label{equation:cf}
  \begin{aligned}[b]
    g\left(x, c\right) &= S_c - S_x \\
      &=
      \begin{aligned}[t]
      &\left(\beta_0 + f_1\left(c_1\right) + \cdots + f_k\left(c_k\right) + \cdots + f_{i,j}\left(c_i, c_j\right)\right) -\\
      &\left(\beta_0 + f_1\left(x_1\right) + \cdots + f_k\left(x_k\right) + \cdots + f_{i,j}\left(x_i, x_j\right)\right)
      \end{aligned}\\
      &=
      \begin{aligned}[t]
      &\left(f_1\left(c_1\right) - f_1\left(x_1\right)\right) + \cdots + \left(f_k\left(c_k\right) - f_k\left(x_k\right)\right) + \cdots +\\
      &\left(f_{i,j}\left(c_i, c_j\right) - f_{i,j}\left(x_i, x_j\right)\right)
      \end{aligned}\\
      &= g\left(x_1, c_2\right) + \cdots + g\left(x_k, c_k\right) + \cdots + g\left(x_i, x_j, c_i, c_j\right)
  \end{aligned}
\end{equation}

\noindent We define the local score gain $g\left(x_i, c_i\right) = f_i\left(c_i\right) - f_i\left(x_i\right)$ as the shape function value difference of changing the main feature $x_i$ to $c_i$.
Similarly, we define the local score gain of a pair-wise interaction term as $g\left(x_i, x_j, c_i, c_j\right) = f_{ij}\left(c_i, c_j\right) - f_{ij}\left(x_i, x_j\right)$.
Then, we can see that the counterfactual constraint $g\left(x, c\right) \geq -S_x$ or $g\left(x, c\right) < -S_x$ is just a linear constraint that consists of a linear combination of shape function value differences.

\subsection{Proximity Requirement}
\label{sec:milp:proximity}

To provide helpful recourse to end users, we want CF examples to be actionable.
One of the most critical measurements of recourse actionability is high proximity between the CF example and the original input, where we want the CF example to only make minimal changes to the original input values~\cite{wachterCounterfactualExplanationsOpening2017,ustunActionableRecourseLinear2019}.
For example, a CF example that suggests increasing annual income by \$5k would be more actionable than another CF example suggesting to increase annual income by \$10k.
We can formulate this proximity requirement as to minimize the distance $d\left(x, c\right)$ between the original input and the CF example---sum of the distances across all features.
\begin{equation}
  \label{equation:distance}
  d\left(x, c\right) = d\left(x_1, c_1\right) + d\left(x_2, c_2\right) + \cdots + d\left(x_k, c_k\right)
\end{equation}

Note that there is no distance cost for pair-wise interaction terms after considering the main effects.
We will discuss our choice of distance functions for continuous and categorical features in-depth in \autoref{sec:practical:distance}.
If all distance functions are linear, or we can pre-compute each $d\left(x_k, c_k\right)$, then the proximity requirement can be formulated as a linear objective function that we want to minimize.

\subsection{Integer Linear Optimization}
\label{sec:milp:ip}

As a gradient-boost tree model, EBM applies equal-frequency binning on continuous features to speed up the training process with a minimal accuracy cost.
For categorical features, EBM uses the discrete levels as bins.
For pair-wise interaction terms, EBM also bins two feature values to construct a lookup table.
Therefore, a CF example can alter the model output if and only if it changes the active bins that some feature values are in.
There are finite number of bins, where each bin provides a local score gain $g\left(x_i, c_i\right)$ and has a distance cost $d\left(x_i, c_i\right)$.
Therefore, generating CF examples for EBM can be thought as solving a variation of \textit{Knapsack Problems}~\cite{salkinKnapsackProblemSurvey1975}.
A knapsack problem considers a set of items where each item has a reward and a weight, and the goal is to find the optimal way to pack items to maximize the total reward under a weight budget.
Popular methods used to solve knapsack problems include integer programming (IP) and dynamic programming.
\tool{} uses IP because (1) it allows users to easily customize optimization constraints~(\autoref{sec:practical:constraint}); (2) users can generate multiple optimal and sub-optimal CF example as recourse~(\autoref{sec:practical:constraint}); (3) modern IP solvers can quickly find a globally optimal solution~(\autoref{sec:practical:speed}).

We express the \tool{} CF generation method as an integer linear programming of the form:

\begin{subequations}
  \label{equation:ip}
  \begin{flalign}
  \min \phantom{.} & \textnormal{distance} \\
  \textnormal{s.t.} \phantom{.} & \textnormal{distance} = \sum_{i=1}^{k} \sum_{b\in{B_i}} d_{ib} v_{ib} & \label{ip:distance} \\
  & -S_x \leq \sum_{i=1}^{k} \sum_{b\in{B_i}} g_{ib} v_{ib} + \sum_{\left(i, j\right) \in N} \sum_{b_1 \in B_i} \sum_{b_2 \in B_j} h_{ijb_1b_2} z_{ijb_1b_2} & \label{ip:cf} \\
  & z_{ijb_1b_2} = v_{ib_1} v_{jb_2} \quad \textnormal{for } \left(i, j\right) \in N, \enskip b_1 \in B_i, \enskip b_2 \in B_j & \label{ip:interaction}\\
  & \sum_{b\in{B_i}}^{} v_{ib} \leq 1 \qquad\quad\! \textnormal{for } i = 1, \dots, k & \label{ip:one}\\
  & v_{ib} \in \left\{0, 1\right\} \qquad\quad\, \textnormal{for } i = 1, \dots, k, \enskip b \in B_i  & \label{ip:binaryv}\\
  & z_{ijb_1b_2} \in \left\{0, 1\right\} \quad\quad\!\! \textnormal{for } \left(i, j\right) \in N, \enskip b_1 \in B_i, \enskip b_2 \in B_j & \label{ip:binaryz}
  \end{flalign}
\end{subequations}
\\
Here, we use an indicator variable $v_{ib}$~(\ref{ip:binaryv}) to denote if a main effect bin is active.
If $v_{ib}=1$, it means that we change the feature value of $x_i$ to the closest value in its bin $b$.
All bin options of $x_i$ are listed in a set $B_i$.
For each feature $x_i$, there can be at most one active bin~(\ref{ip:one}); if there is no active bin, then we do not change the feature value of $x_i$.
Similarly, we use an indicator variable $z_{ijb_1b_2}$~(\ref{ip:binaryz}) to denote if an interaction effect is active.
This interaction effect is active if and only if bin $b_1$ of feature $x_i$ and bin $b_2$ of feature $x_j$ are both active~(\ref{ip:interaction}).
$N$ denotes a set of feature pairs that the given EBM computes interaction effects from.
Constraint~(\ref{ip:distance}) determines the total distance cost for a potential CF example; it uses a set of pre-computed distance costs $d_{ib}$ of changing one feature $x_i$ to the closest value in bin $b$~(\autoref{sec:milp:proximity}).

Constraint~(\ref{ip:cf}) ensures that any solution would flip the prediction of the given EBM model~(\autoref{sec:milp:cf}).
Constraint~(\ref{ip:cf}) is used when the model predicts the original input as negative; if the original prediction is positive, we only need to change $\leq$ to $>$ (\autoref{sec:milp:cf}).
Here, $g_{ib}$ and $h_{ijb_1b_2}$ denote pre-computed local score gains of activating bin $b$ in $x_i$ and activating the interaction effect $z_{ijb_1b_2}$, respectively.
Note that activating one bin can trigger multiple interaction effects, but $h_{ijb_1b_2}$ is only counted when both $v_{ib_1}$ and $v_{jb_2}$ are active~(\ref{ip:cf} and \ref{ip:binaryz}).
Therefore, we compute $g_{ib}$ by preemptively adding the shape function differences of all \textcolor[HTML]{E03177}{partially affected interaction effects} to the shape function difference of the main effect.
For example, if $N = \left\{\left(i, j\right), \left(i, m\right), \left(l, m\right)\right\}$, we compute $g_{ib}$ and $g_{jb}$ as:
\begin{subequations}
  \label{equation:main-score-gain}
\begin{align}
g_{ib} = &
\left(f_i\left(x_{ib}\right) - f_i\left(x_{i0}\right)\right) +
  \left(\textcolor[HTML]{E03177}{f_{ij}\left(x_{ib}, x_{j0}\right) - f_{ij}\left(x_{i0}, x_{j0}\right)}\right) + & \nonumber\\
&\left(\textcolor[HTML]{E03177}{f_{im}\left(x_{ib}, x_{m0}\right) - f_{im}\left(x_{i0}, x_{m0}\right)}\right)  & \label{equation:main-score-gain:i}\\
g_{jb} = &\left(f_j\left(x_{jb}\right) - f_j\left(x_{j0}\right)\right) +
  \left(\textcolor[HTML]{E03177}{f_{ij}\left(x_{i0}, x_{jb}\right) - f_{ij}\left(x_{i0}, x_{j0}\right)}\right) & \label{equation:main-score-gain:j}
\end{align}
\end{subequations}

Here, $x_{ib}$ denotes the closest value of bin $b$ of feature $x_i$, and $x_{i0}$ denotes the original value of feature $x_i$.
In \ref{equation:main-score-gain:i}, we add two \textcolor[HTML]{E03177}{partial interaction score gains} because activating bin $b$ of feature $x_i$ affects two interaction terms $\left(i, j\right)$ and $\left(i, m\right)$.
Similarly, \ref{equation:main-score-gain:i} only includes one \textcolor[HTML]{E03177}{partial interaction score gain} because activating bin $b$ of feature $x_j$ only affects one interaction term $\left(i, j\right)$.

However, when both $v_{ib_1}$ and $v_{jb_2}$ are active, the interaction score gain should be $f_{ij}\left(x_{ib_1}, x_{jb_2}\right) - f_{ij}\left(x_{i0}, x_{j0}\right)$.
Therefore, we need to offset two \textcolor[HTML]{E03177}{partial interaction score gains} added preemptively when computing $g_{ib}$ and $g_{jb}$~(\ref{equation:main-score-gain:i} and \ref{equation:main-score-gain:j}).
To do that, we simply subtract \textcolor[HTML]{E03177}{them} when computing the interaction score gain $h_{ijb_1b_2}$:

\begin{equation}
  \begin{aligned}[c]
    \label{equation:offset}
    h_{ijb_1b_2} = &\left(f_{ij}\left(x_{ib_1}, x_{jb_2}\right) -  f_{ij}\left(x_{i0}, x_{j0}\right)\right) -\\
      &\left(\textcolor[HTML]{E03177}{f_{ij}\left(x_{ib_1}, x_{j0}\right) -  f_{ij}\left(x_{i0}, x_{j0}\right)}\right) -\\
      &\left(\textcolor[HTML]{E03177}{f_{ij}\left(x_{i0}, x_{jb_2}\right) -  f_{ij}\left(x_{i0}, x_{j0}\right)}\right)
  \end{aligned}
\end{equation}
\\
Once trained, the EBM model transforms all parameters into lookup histograms and lookup tables~(\autoref{sec:method:model}), so we can quickly pre-compute all $g_{ib}$ and $h_{ijb_1b_2}$ terms.
Furthermore, we can linearize the binary variable multiplication constraint~(\ref{ip:interaction}) as three linear constraints:
(1) $z_{ijab} \leq v_{ia}$; (2) $z_{ijab} \leq v_{jb}$; (3) $z_{ijab} \geq v_{ia} + v_{jb} - 1$.
Then, all constraints (\ref{ip:distance}--\ref{ip:binaryz}) are linear, and (\ref{equation:ip}) is an integer linear program with all binary variables, which can be efficiently solved by modern IP solvers~\cite{saltzmanCoinOrOpenSourceLibrary2002}.
As this formulation considers all possible effective changes to the original input, the solution to (\ref{equation:ip}) is guaranteed to be the optimal CF example regarding the given distance functions.

\subsection{Choice of Distance Function}
\label{sec:practical:distance}

It is challenging to define a distance function that can accurately measure the difficulty for end users to change a feature~\cite{barocasHiddenAssumptionsCounterfactual2020}.
In \tool{}, we use the $\ell_1$ distance to measure the distance between the original input and the CF example across continuous features.
As different continuous features often have different scales, we divide each feature-wise distance by the median absolute deviation (MAD) of that feature on the training set, which is a common choice among other CF generation methods~\cite[e.g.,][]{kanamoriDACEDistributionAwareCounterfactual2020,mothilalExplainingMachineLearning2020,wachterCounterfactualExplanationsOpening2017}.
MAD provides a robust way to measure the variance within each feature.
Here, $n$ is the size of the training set. Dividing the $\ell_1$ distance with MAD implies that it is relatively easier for end users to change a high-variance features than low-variance features.
\begin{equation}
  \label{equation:cont-distance}
  d_{\text{cont}}\left(x_i, c_i\right) = \frac{\left| x_i - c_i \right|}{\text{Median}_{j = 1}^{n}\left(\left| x_i^{\left(j\right)} - \text{Median}_{p = 1}^{n}\left(x_{i}^{\left(p\right)}\right) \right| \right)}
\end{equation}

It is harder to define the distance for categorical features.
Some CF methods use $1$ for features having the same level and $0$ for different level~\cite{mothilalExplainingMachineLearning2020}, and others consider the probability that two examples would share the same level~\cite{wexlerWhatIfToolInteractive2019}.
In \tool{}, we use the complement of the probability of seeing one level based on its frequency in the training set.
Here, $n$ is the size of the training set and $\mathbb{I}$ is the indicator function.
This distance definition implies that it is easier for end users to change to a more frequent level in a given categorical feature.
\begin{equation}
  \label{equation:cat-distance}
  d_{\text{cat}}\left(x_i, c_i\right) = 1 - \frac{\sum_{j=1}^{n}\mathbb{I}\left({x_i^{\left(j\right)} = c_i }\right)}{n}
\end{equation}

After counting distance costs of all bins of main effects, we re-weight distance costs of all categorical bins so that the average of continuous feature distances is the same as the average of categorical feature distances.
There is no right way to choose distance functions~\cite{mothilalExplainingMachineLearning2020,barocasHiddenAssumptionsCounterfactual2020}.
Fortunately, all distances are pre-computed before solving the actual IP, and \tool{} provides flexible APIs to let developers use their own distance functions.

\setlength{\columnsep}{10pt}%
\setlength{\intextsep}{0pt}%
\begin{wrapfigure}{R}{0.19\textwidth}
  \vspace{0pt}
  \centering
  \includegraphics[width=0.18\textwidth]{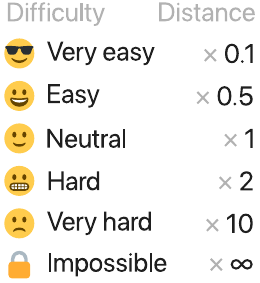}
  \vspace{-10pt}
  \caption[]{
    Distance multipliers of difficulties.
  }
  \Description{
    A table mapping difficulty icons to their distance multipliers.
    Very easy maps to ``⨉ 0.1''.
    Easy maps to ``⨉ 0.5''.
    Neutral maps to ``⨉ 1''.
    Hard maps to ``⨉ 2''.
    Very hard maps to ``⨉ 10''.
    Impossible maps to ``⨉ infinity''.
  }
  \label{fig:appendix-diff-map}
\end{wrapfigure}
Ultimately, we believe that instead of researchers searching for a one-fit-all distance function, we should enable end users to directly specify their own difficulty to change features~(\aptLtoX[graphic=no,type=html]{\textbf{G2}}{\ref{item:g2}}).
To do that, \tool{} provides end users with an interface to select feature \textit{difficulties} by clicking buttons~(\autoref{fig:scenario}\figpart{-B1}).
Internally, \tool{} assigns each difficulty level with a constant multiplier~(\autoref{fig:appendix-diff-map}).
Before solving the IP, the tool multiplies the pre-computed distances of all bins in a feature with this constant multiplier.
For example, if a user selects ``very easy'' for feature $i$, then the distance between the original value $c_i$ and the closest value in bin $b_{ij}$ of feature $i$ is computed as $0.1 \times d\left(b_{ij}, c_i\right)$.
If a user selects the ``impossible to change'' difficulty, \tool{} will remove all variables associated with this feature in the IP.
Therefore, when generating new recourse plans, \tool{} would prioritize features that are easier to change and would not consider features that are impossible to change.
We choose six levels of feature difficulties because we observe that we can mix and match these six levels on different features to flexibly fine-tune recourse generation in our experiments with six datasets.
We choose the four constant multipliers $[0.1, 0.5, 2, 10]$ because they can noticeably affect the IP solutions with ``appropriate'' strengths.
However, researchers and developers can easily change these constant values and also the difficulty granularity (e.g., with only three levels ``very easy'', ``neutral'', and ``impossible'') in their specific use cases.

\vspace{-2pt} %
\subsection{Generalization to Regression}
\label{sec:practical:regression}

\citet{barocasHiddenAssumptionsCounterfactual2020} finds that algorithmic recourse literature often assumes the ML model outcome to be binary, such as loan approval, school acceptance, and hiring decision.
However, in reality, end users also need recourse for AI-generated decisions on continuous values such as a loan's interest rate.
\tool{} supports generating CF examples for regression problems.
To do that, we only need to modify the CF constraint to bound the needed score gain to meet the desired range provided by the end user~(\autoref{sec:milp:cf}).
Then, we can update the left side value $-S_x$ and the inequality in \ref{ip:cf} to reflect the score gain boundaries.
This constraint would still be linear, and IP solver can solve the whole program.
For example, to increase the predicted continuous value (e.g., interest rate) by at least $\delta$, we only need to modify \ref{ip:cf} to be:
\vspace{-3pt}
\begin{equation}
  \label{regression:cf}
  \delta \leq \sum_{i=1}^{k} \sum_{b\in{B_i}} g_{ib} v_{ib} + \sum_{\left(i, j\right) \in N} \sum_{b_1 \in B_i} \sum_{b_2 \in B_j} h_{ijb_1b_2} z_{ijb_1b_2}
\end{equation}

\vspace{-4pt} %
\subsection{Generalization to Multiclass Classification}
\label{sec:practical:multiclass}

In addition to regression, our IP can be easily generalized for multiclass classification.
Compared to binary EBM, multiclass EBM~\cite{zhangAxiomaticInterpretabilityMulticlass2019} uses a multiclass cross entropy as its loss function and softmax as its link function.
Once trained, an $n$-class EBM has a similar structure as the binary EBM.
However, there are no interaction terms in a multiclass EBM, and each bin of a feature now has $n$ associated additive scores instead of just $1$ score as in binary EBM.
During inference, the $n$-class EBM adds up the additive scores from all features and an intercept for each class.
For example, we use $S_x^1$ to denote the score for class 1 of input $x$, then $S_x^1 = \beta_0^1 + f_1^1\left(x_1\right) + f_2^1\left(x_2\right) + \cdots + f_k^1\left(x_k\right)$.
Next, the softmax link function~(\autoref{equation:softmax}) rescales $n$ scores $S_x^1, S_x^2, \dots, S_x^n$ to $n$ class probabilities $\sigma_x^1, \sigma_x^2, \dots, \sigma_x^n$, where $\sum_{j=0}^{n}\sigma_x^j = 1$.
Finally, the multiclass EBM chooses the class $j$ with the largest $\sigma_x^j$ as the final prediction.
\vspace{-1pt}
\begin{equation}
  \label{equation:softmax}
  \sigma_x^p = \frac{\exp{\left(S_x^p\right)}}{\sum_{j=1}^{n}\exp{\left(S_x^j\right)}}
\end{equation}
\vspace{-1pt}
Note that the softmax function is monotonic and it preserves the rank order of its input values.
In other words, to make a multiclass EBM predict class $p$ on a CF example $c$, we only need to make $S_c^j < S_c^p$ for $j = 1,\dots,n$ and $j \neq p$, which can be written as $n - 1$ linear constraints.
Therefore, the \tool{} CF generation method for multiclass classification (target class is $p$) can be written as the following integer linear program:
\begin{subequations}
  \label{equation:multi}
  \begin{flalign}
  \min \phantom{.} & \textnormal{distance} \\
  \textnormal{s.t.} \phantom{.} & \textnormal{distance} = \sum_{i=1}^{k} \sum_{b\in{B_i}} d_{ib} v_{ib} & \label{multi:distance} \\
  &S_x^j + \sum_{i=1}^{k} \sum_{b\in{B_i}} g_{ib}^j v_{ib} < S_x^p + \sum_{i=1}^{k} \sum_{b\in{B_i}} g_{ib}^p v_{ib} \nonumber\\
  &\qquad\qquad\qquad\qquad\quad \textnormal{for } j = 1, \dots, n \enspace \textnormal{and} \enspace j \neq p
  & \label{multi:cf}\\
  & \sum_{b\in{B_i}}^{} v_{ib} \leq 1 \qquad\qquad \textnormal{for } i = 1, \dots, k & \label{multi:one}\\
  & v_{ib} \in \left\{0, 1\right\} \qquad\qquad\,\, \textnormal{for } i = 1, \dots, k, \enskip b \in B_i  & \label{multi:binaryv}
  \end{flalign}
\end{subequations}
\\
In constraint~\ref{multi:cf}, $S_x^j$ is the total score for class $j$ of the original input $x$.
Similar to $g_{ib}$ in \ref{ip:cf}, $g_{ib}^j$ denotes the score gain for class $j$ of changing feature $x_i$ to the closest value in its bin $b$.
All constants $S_x^j$ and $g_{ib}^j$ can be pre-computed.

\vspace{-2pt} %
\subsection{Support Various Actionability Constraints}
\label{sec:practical:constraint}

To generate recourses that are actionable for end users, we not only prefer CF examples that are close to the original input~(\autoref{sec:milp:proximity}), but also \textit{concise}~\cite{leGRACEGeneratingConcise2020}, \textit{diverse}~\cite{mothilalExplainingMachineLearning2020,russellEfficientSearchDiverse2019}, and respect to individual end users' \textit{preferences}~\cite{barocasHiddenAssumptionsCounterfactual2020,keaneIfOnlyWe2021}.
With \tool{}, we can generate CF examples with these desired properties by formulating these requirements as linear constraints in the IP.
For example, to generate concise or sparse CF examples---examples that only change a few features from the original input---we can introduce a linear constraint to bound the total number (up to $p$) of active variables for main effects: $\sum_{i=1}^{k}\sum_{b\in{B_i}} v_{ib} \leq p$.
To generate diverse CF examples, we can solve the same IP multiple times, where each time we add a new constraint to force the solver to avoid previous solutions.
For example, we can set $v_{ib_i}v_{jb_j}v_{kb_k} = 0$ for new iterations where $\{v_{ib_i}=1,v_{jb_j}=1,v_{kb_k}=1\}$ is a previous solution.
Since all variables are binary, we can linearize these multiplication constraints~\cite{gloverImprovedLinearInteger1975}.
With this approach, the generated $k$ diverse solutions are also guaranteed to be the top-$k$ optimal solutions.
Similarly, if we have prior knowledge of end users' preferences, such as difficulties and actionable ranges of individual features, we can adjust the distance costs during the pre-computation process.
Therefore, the flexibility of IP helps us operationalize the design of \tool{}~(\aptLtoX[graphic=no,type=html]{\textbf{G2}}{\ref{item:g2}}).

\vspace{-2pt} %
\subsection{CF Generation Method Comparison}
\label{sec:practical:compare}

Our CF generation method is the first and only CF algorithm specifically developed for EBM models.
Before our method, ML researchers and developers would need to use model-agnostic algorithms like genetic algorithm~\cite{schleichGeCoQualityCounterfactual2021} and KD-tree~\cite{vanlooverenInterpretableCounterfactualExplanations2020} to generate recourse plans for EBM models.
Our technique is guaranteed to outperform or tie with these algorithms if we measure the quality of CFs by their distances (e.g., $\ell_1$ distance) to the original input.
This is because our technique formulates CF generation as a linear optimization program~(\autoref{sec:milp:ip}) that minimizes the distance between the modified and original inputs.
For completeness, we have included such comparison results in \autoref{tab:comparison} to give readers a sense of how far from optimal existing CF generation methods are in terms of distance.

\begin{table*}[tb]\centering
  \ra{1.3}
  \caption[]{
    We compare our method with two existing CF generation methods: genetic algorithm and KD-tree.
    We train three EBM binary classifiers on LendingClub, German Credit, and Adult datasets, and then apply three CF algorithms to generate CFs for test samples that are rejected for a loan.
    The results highlight that our method significantly outperforms existing methods.
    In particular, CFs generated by our method are \textit{closer} to the original input, \textit{more sparse}, and encounter \textit{less failures}.}
  \label{tab:comparison}
  \begin{tabular}{lrrrrrr}
    \toprule
    & \phantom{xx} & Mean Distance & \phantom{x} & Mean Number of Features Changed & \phantom{x} & Number of Failures\\
    \midrule

    \multicolumn{7}{c}{Lending Club (378 samples)}\\

    Our Method && \textbf{0.1836} && \textbf{2.2222} && \textbf{0 }\\
    Genetic Algorithm~\cite{schleichGeCoQualityCounterfactual2021} && 3.1950 && 10.2520 && 1 \\
    KD Tree~\cite{vanlooverenInterpretableCounterfactualExplanations2020} && 3.7388 && 10.8360 && 6 \\
    \arrayrulecolor{tablegray}
    \cmidrule{1-7}

    \multicolumn{7}{c}{German Credit (239 samples)}\\
    Our Method && \textbf{1.1392} && \textbf{2.0962} && \textbf{0} \\
    Genetic Algorithm~\cite{schleichGeCoQualityCounterfactual2021} && 6.8573 && 9.3305 && 0 \\
    KD Tree~\cite{vanlooverenInterpretableCounterfactualExplanations2020} && 7.3565 && 9.9414 && 0 \\

    \cmidrule{1-7}

    \multicolumn{7}{c}{Adult (400 samples)}\\

    Our Method && \textbf{1.6856} && \textbf{2.4075} && \textbf{0} \\
    Genetic Algorithm~\cite{schleichGeCoQualityCounterfactual2021} && 4.9231 && 4.6475 && 0 \\
    KD Tree~\cite{vanlooverenInterpretableCounterfactualExplanations2020} && 5.1082 && 4.9500 && 0 \\

    \bottomrule
  \end{tabular}
  \Description{
    A table of the counterfactual explanation quality on different datasets with different generation methods.
    On the lending club dataset, our method outperforms the Genetic Algorithm and KD Tree with a mean distance of 0.1836 (vs. 3.195 and 3.7388), the mean number of features changed of 2.2222 (vs. 10.252 and 10.8360), and the number of failures of 0 (vs. 1 and 6).
    On the german credit dataset, our method outperforms the Genetic Algorithm and KD Tree with a mean distance of 1.1392 (vs. 6.8573 and 7.3565), the mean number of features changed of 2.0962 (vs. 9.3305 and 9.9414), and the number of failures of 0 (vs. 0 and 0).
    On the adult dataset, our method outperforms the Genetic Algorithm and KD Tree with a mean distance of 1.6856 (vs. 4.9231 and 5.1082), the mean number of features changed of 2.4075 (vs. 4.6475 and 4.95), and the number of failures of 0 (vs. 0 and 0).
  }
\end{table*}

In the comparison experiment, we train three EBM binary classifiers on LendingClub~\cite{LendingClubOnline2018}, Adult~\cite{kohaviScalingAccuracyNaivebayes1996}, and German Credit~\cite{duaUCIMachineLearning2017} datasets.
We use our IP approach, genetic algorithm, and KD-tree to generate CFs for test samples that are rejected for a loan (378, 400, and 239 samples from three datasets).
We use the DICE library's implementation~\cite{mothilalExplainingMachineLearning2020} of the genetic algorithm and KD-tree.
We disable our method's default categorical distance~(\autoref{sec:practical:distance}) to match the other two algorithms (distance is $1$ if the category is changed and 0 otherwise).
All three algorithms use MAD adjusted $\ell_1$ to measure the distance of continuous variables.
The distance between two samples is defined as the mean of all categorical and continuous distances.
The results~(\autoref{tab:comparison}) highlight that compared to existing methods, CFs generated by our method are significantly \textit{closer} to the original input, \textit{more sparse}, and encounter \textit{fewer failures}.

\subsection{Fast CF Generation}
\label{sec:practical:speed}

In many cases of providing algorithmic recourse, we need to prioritize CF example generation speed over the optimality of generated CF examples~\cite{schleichGeCoQualityCounterfactual2021}.
With \tool{}, modern IP solvers can efficiently solve the program~(\aptLtoX[graphic=no,type=html]{Equation \ref{equation:ip}}{\autoref{equation:ip}}).
The complexity of solving an integer linear program increases along two factors: the number of variables and the number of constraints.
Here, all variables are binary---making the program easier to solve than a program with non-binary integer variables.
For any dataset, there are always exactly 3 constraints from \ref{ip:distance}, \ref{ip:cf}, and \ref{ip:one}.
The number of constraints from \ref{ip:interaction} increases along the number of interaction terms $|N|$ and the number of bins per feature $|B_i|$ on these interaction terms.
In practice, $|N|$ and $|B_i|$ are often bounded to ensure GAMs are interpretable.
For example, by default the popular GAM library InterpretML~\cite{noriInterpretMLUnifiedFramework2019} bounds $|N| \leq 10$ and $|B_i| \leq 32$.
Therefore, in the worst-case scenario with 10 continuous-continuous interaction terms, there will be at most $10 \times 32 \times 32 = 10,240$ constraints from \ref{ip:interaction}.
For example, on the Communities and Crime dataset~\cite{redmondDatadrivenSoftwareTool2002} with 119 continuous features, 1 categorical feature, and 10 pairwise interaction terms, there are about 7.2k constraints and 3.6k variables in our program.
It only takes about 0.5--3.0 seconds to generate a recourse plan using Firefox Browser on a MacBook.

In addition, in applications where the generation speed is critical, developers can significantly improve the run time by filtering less effective bins during the pre-computation process, which decreases the number of variables quadratically.
First, developers can filter out main effect bins that give opposite score gains from the objective (i.e., positive score gain when the goal is to lower the prediction score).
By default, \tool{} does not apply this filtering, because in rare cases the score gains of associated interaction terms can offset the opposite score gain from the main effect.
By filtering out bins with opposite score gains, \tool{} can consistently generate CF examples in under 1 second in end users' browsers~(\autoref{sec:ui}).
To further improve the speed, developers can also filter out main effect bins that give similar score gains as existing bins but have a higher distance cost.

\clearpage{}
\section{Supplementary Figures}

\begin{figure*}[b!]
  \includegraphics[width=\linewidth]{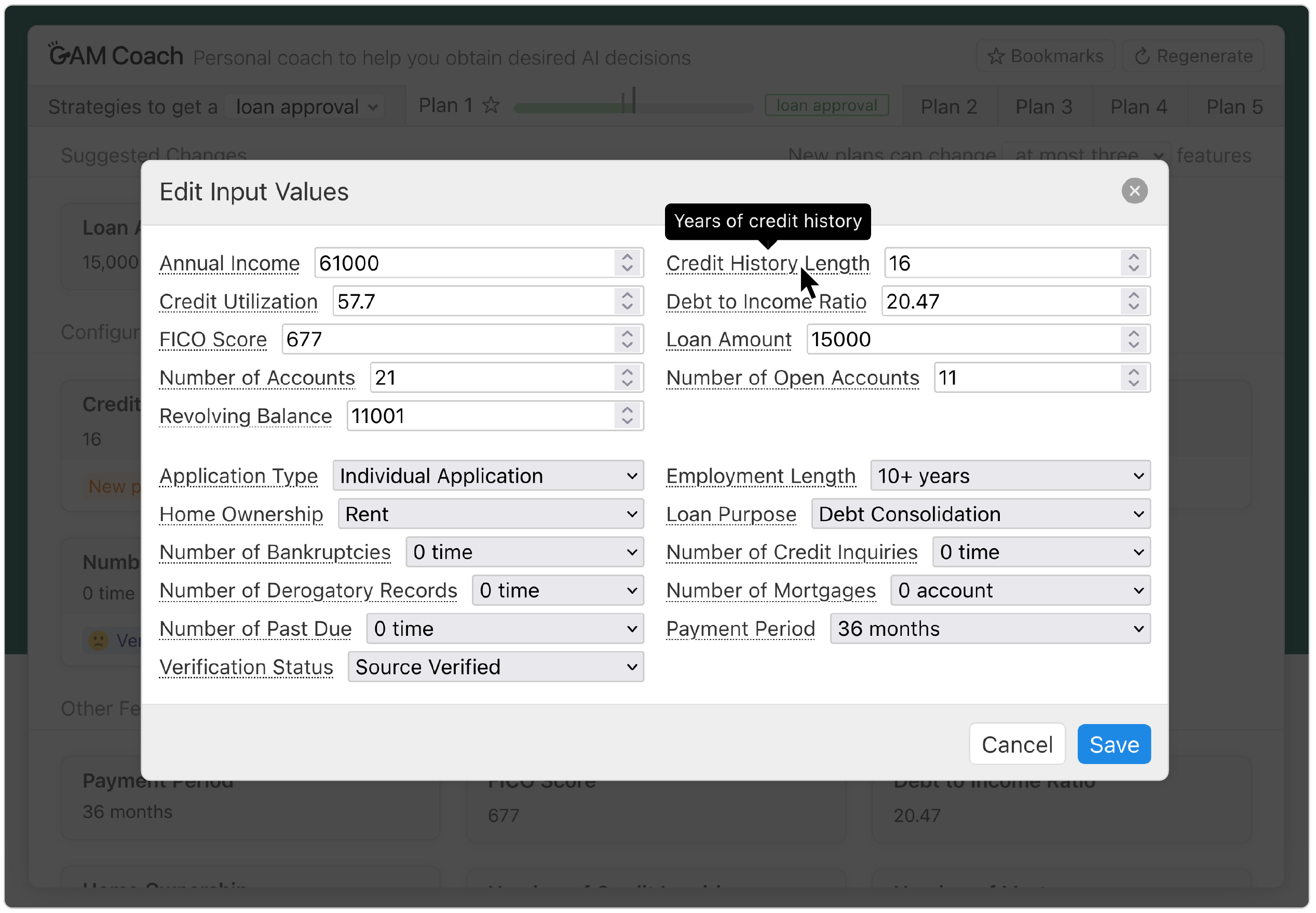}
  \captionof{figure}[]{
    To help user study participants imagine the loan application scenario, the \tool{} interface allows participants to change the input values of the hypothetical loan applicant.
    The top row includes input fields for the 9 continuous features, and the bottom row contains dropdowns for the 11 categorical features used in the EBM model.
    Users can hover over the feature name to see the detailed description for that feature.
  }
  \Description{
    A screenshot of a pop-up window where users can change the input values.
    On top of the window, users can use text fields to enter new values for continuous features.
    On the bottom, users can use drop-down menus to change the levels of categorical features.
  }
  \label{fig:appendix-input}
\end{figure*} 
\end{document}